\documentclass[10pt,journal,compsoc]{IEEEtran}

\usepackage{amsmath,amssymb,amsfonts, mathrsfs}
\usepackage{graphicx, subfigure, multirow}
\usepackage{algorithm, algorithmic}
\usepackage{cite}
\usepackage{xcolor}
\usepackage{setspace}
\usepackage{subfigure}
\usepackage{colortbl}
\usepackage{enumitem}

\begin{document}
\title{Generative Adversarial Networks Unlearning}

\author{Hui Sun,
        Tianqing Zhu,~\IEEEmembership{Member,~IEEE,}
        Wenhan Chang,
        and Wanlei Zhou,~\IEEEmembership{Senior Member,~IEEE}% <-this % stops a space
\IEEEcompsocitemizethanks{\IEEEcompsocthanksitem Hui Sun and Wenhan Chang are with the China University of Geosciences, Wuhan, China; Tianqing Zhu is with the University of Technology Sydney; Wanlei Zhou is with the City University of Macau.\protect\\
\IEEEcompsocthanksitem Tianqing Zhu is the corresponding author. E-mail: tianqing.zhu@ieee.org}% <-this % stops a space
}


\markboth{Journal of \LaTeX\ Class Files,~Vol.~14, No.~8, August~2015}%
{Shell \MakeLowercase{\textit{et al.}}: Bare Advanced Demo of IEEEtran.cls for IEEE Computer Society Journals}



\IEEEtitleabstractindextext{%
\begin{abstract}

As machine learning continues to develop, and data misuse scandals become more prevalent, individuals are becoming increasingly concerned about their personal information and are advocating for the right to remove their data. Machine unlearning has emerged as a solution to erase training data from trained machine learning models. Despite its success in classifiers, research on Generative Adversarial Networks (GANs) is limited due to their unique architecture, including a generator and a discriminator. One challenge pertains to generator unlearning, as the process could potentially disrupt the continuity and completeness of the latent space. This disruption might consequently diminish the model's effectiveness after unlearning. Another challenge is how to define a criterion that the discriminator should perform for the unlearning images.
In this paper, we introduce a substitution mechanism and define a fake label to effectively mitigate these challenges. 
Based on the substitution mechanism and fake label, we propose a cascaded unlearning approach for both item and class unlearning within GAN models, in which the unlearning and learning processes run in a cascaded manner.
We conducted a comprehensive evaluation of the cascaded unlearning technique using the MNIST and CIFAR-$10$ datasets, analyzing its performance across four key aspects: unlearning effectiveness, intrinsic model performance, impact on downstream tasks, and unlearning efficiency.
Experimental results demonstrate that this approach achieves significantly improved item and class unlearning efficiency, reducing the required time by up to $185\times$ and $284\times$ for the MNIST and CIFAR-$10$ datasets, respectively, in comparison to retraining from scratch. Notably, although the model's performance experiences minor degradation after unlearning, this reduction is negligible when dealing with a minimal number of images (e.g., $64$) and has no adverse effects on downstream tasks such as classification.

\end{abstract}

}

\maketitle

% is not selected <OR> if conference mode is selected - because compsoc
% conference papers position the abstract like regular (non-compsoc)
\IEEEdisplaynontitleabstractindextext

% \fi
\IEEEpeerreviewmaketitle

\ifCLASSOPTIONcompsoc
\IEEEraisesectionheading{\section{Introduction}\label{sec:introduction}}
\else
% \section{Introduction}
% \label{sec:introduction}
\fi

In the era of big data, people generate vast volumes of data through a variety of activities, such as sharing, communicating, shopping, and more. These data have enabled significant advancements in deep learning. However, the utilization of machine learning models has given rise to privacy concerns \cite{10.1145/3547330,9627776}. For instance, adversaries can potentially infer that a specific individual has received a hospital diagnosis by querying machine learning models. This is the so-called membership inference attack \cite{Shokri7958568, 9723588}, deducing whether an image has been used for model training. Owing to the concern of privacy leakage, people ought to be capable to withdraw their data. The European Union has enacted the General Data Protection Regulation (GDPR) \cite{GuoJiaShuJu}, which emphasizes the right to be forgotten. As per this regulation, organizations are required to erase any personal data from their records, including from pre-trained machine learning models, upon request.

\begin{figure}[tbp]
\centering
\includegraphics[scale=0.65]{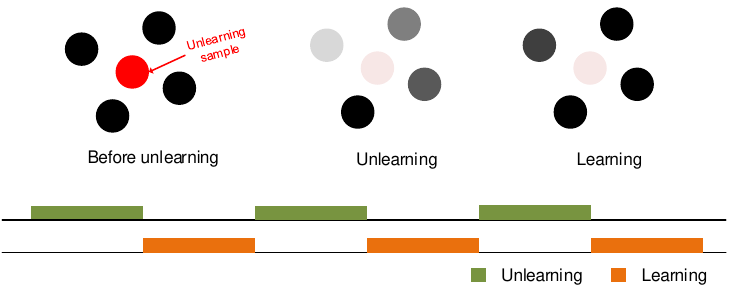}
\caption{A cascaded unlearning algorithm. The unlearning process can potentially compel the target GAN model to unlearn images that should be retained. To address the risk of over-unlearning, our cascaded unlearning algorithm incorporates both an unlearning and a learning algorithm, operating in a cascaded manner to ensure optimal results.}
\label{Fig: cascaded unlearning algorithm}
\end{figure}

Withdrawing data from a pre-trained machine learning model is referred to as "machine unlearning" \cite{10.1145/3603620, 10189868}. This process aims to eliminate the influence of specific images or a certain class from a pre-trained model, namely, "item unlearning" and "class unlearning." 
The exact unlearning is to achieve complete data withdrawal while maintaining model fidelity. This involves retraining the model using the remaining data. However, this approach is associated with significant time and computing resource requirements.
A potential solution is presented by the SISA framework \cite{DBLP:conf/sp/BourtouleCCJTZL21}. In this framework, data is partitioned into shards and slices, and an intermediate model is trained for each shard. The final output is obtained through the aggregation of multiple models across these shards. When unlearning is required, partial intermediate models are retrained based on specific requests. This approach significantly reduces computing resource consumption but comes at the cost of substantial storage space usage.
Modifying the final model parameters is a popular alternative \cite{9844865learntoforget,DBLP:journals/corr/abs-2201-05629, DBLP:conf/ndss/WarneckePWR23}. This method also requires model retraining, but it utilizes an unlearning loss function. In essence, the unlearning model approximates the model that has been retrained using the remaining data. 
Notably, these methods generally involve the remaining training images. For example, the SISA framework is built upon the foundation of these remaining training images, and modifying the final model parameters utilizes these remaining training images to mitigate the risk of over-unlearning. Over-unlearning implies that the model becomes unable to handle the tasks associated with the remaining data.
However, the inclusion of remaining training images raises concerns about privacy leakage, particularly for sensitive data such as medical records and facial images. Unlearning with few or no remaining training images has become increasingly urgent and important. This is commonly referred to as "few-shot" and "zero-shot" unlearning, respectively.
% It's important to note that these techniques may potentially lead to a potential over-unlearning, which implies that the model becomes unable to handle the tasks associated with the remaining data.
%To mitigate the risk of over-unlearning, it is essential to incorporate positive information, which can be derived from the remaining data or even from the model itself. 
Furthermore, it's worth noting that existing research primarily focuses on classification models, and there is a limited exploration of these concepts in the context of generative models, such as Generative Adversarial Networks (GAN), due to their unique structure.

Popular unlearning methodologies cannot be directly applied to a GAN model due to GAN's unique architecture. A GAN model consists of a generator and a discriminator that the discriminator learns from the training data thus guiding the generator to generate plausible images. If we construct the SISA framework on a GAN model, we will attain multiple generators as well as multiple images upon a request, which is impractical. Therefore, we can only modify the final model parameters, but this method also meets two critical challenges. 

The first challenge pertains to the generator, as unlearning potentially jeopardizes the latent space's continuity. Latent space continuity denotes that proximate points within the latent space should not yield entirely disparate images. Consequently, direct withdrawal of data results in the discontinuity of the latent space, ultimately compromising the subsequent generation performance.
The second challenge pertains to the discriminator: how to define a criterion that the discriminator should perform for the unlearning images. In the case of classifier unlearning, the criterion utilizes the outputs from test images. The classifier exhibits distinct behavior between training and test images, and the disparity has been associated with the success of membership inference techniques \cite{Shokri7958568, 10.1145/3523273}. In essence, if the classifier demonstrates equivalent performance on both unlearning and test images, it indicates a substantial reduction in the classifier's confidence for these unlearning images, thereby completing the unlearning.
However, we have observed that the discriminator's outputs for training and test images exhibit less pronounced differences. For instance, consider the case of StyleGAN2 shown in Figure \ref{Fig: mnist_more_D_scores}. The intertwined curves representing training and test images suggest that adopting the outputs of test images as the criterion might not yield significant insights for effective unlearning. This observation raises concerns regarding the potential efficacy of the unlearning process. Prior to the development of the unlearning algorithm, addressing these challenges becomes imperative.

To tackle the first challenge, we introduce a substitute mechanism that presents an alternative image to replace the image undergoing the unlearning process. This alternative image facilitates an alternative latent-image mapping, which would keep the continuity of the latent space. The substitute mechanism provides separate solutions for item and class unlearning, which are detailed in section \ref{sec:substitute mechanism}. For the second challenge, we define a fake label as the fixed criterion for the discriminator unlearning. This fake label is significantly smaller than the discriminator's output scores for training images. When unlearning an image, the output of the discriminator is expected to approximate the fake label. This would result in a significant decrease in the discriminator's confidence when confronted with the unlearning image.
%According to the experimental results, the transition value between the outputs of test images and generated images, i.e., $0.1$ makes a successful unlearning.

This paper proposes a cascaded unlearning algorithm in which the unlearning and learning processes operate in a cascaded manner, addressing both item unlearning and class unlearning. The unlearning phase tries to unlearn specific images or a class, with the help of the substitute mechanism and defined fake label. While the learning phase utilizes learning information to prevent over-unlearning, which stems from a few training images in the few-shot case or the raw GAN model in the zero-shot case. 
Our cascaded unlearning stands out not only for its effectiveness in facilitating unlearning, but also for its efficiency and privacy preservation.

An overview of contributions. Our contributions to the
current work include the following:
\begin{enumerate}[leftmargin=*]
    \item The first work to the analysis of challenges that GAN unlearning meets.
    \item A systematic investigation of our proposed cascaded unlearning, covering both few-shot and zero-shot settings.
    \item The first machine unlearning of GAN models, including item and class unlearning.
    \item A comprehensive evaluation framework, including unlearning effectiveness, the model performance, the impact on downstream tasks such as classification, and unlearning efficiency.
\end{enumerate}

\section{Related Work}

\subsection{Machine Unlearning}

Machine unlearning supports the "right to be forgotten," a concept introduced by the European Union's General Data Protection Regulation (GDPR) \cite{GuoJiaShuJu}. In the machine learning field, machine unlearning methods remove certain data from the data center as well as from the derived-trained models. 

The direct way for unlearning is to retrain the model on the remaining data, ensuring both absolute withdrawal and model fidelity. Retraining undeniably achieves precise unlearning \cite{nguyenSurveyMachineUnlearning2022}, but it inherently comes with substantial computational costs, particularly for larger models \cite{UnrollingSGDThudiDCP22}.
Approximate unlearning techniques such as SISA framework \cite{DBLP:conf/sp/BourtouleCCJTZL21} enhance the efficiency of unlearning. This framework entails partitioning data into shards and slices, followed by the training of an intermediate model for each shard. The ultimate outcome is an aggregation of multiple models derived from these shards. During the unlearning process, the model owner retrieves the specific unlearning data and initiates retraining from the relevant left intermediate models. This strategic approach significantly conserves computational resources
A more efficient and prevalent strategy is to modify the final model parameters \cite{DBLP:conf/aaai/GravesNG21,DBLP:conf/icml/GuoGHM20,DBLP:conf/nips/SekhariAKS21} via gradient descent applied to unlearning data. This approach asserts that the unlearnt model maintains uniform performance for both the unlearnt and non-member data. Simultaneously, training data is employed to mitigate the potential for catastrophic unlearning. Given the sensitive nature of training data, research has placed considerable emphasis on scenarios where a limited number of or even no training images are involved. Specifically, these cases are known as few-shot \cite{DBLP:journals/corr/abs-2107-03860, DBLP:journals/corr/abs-2205-15567} and zero-shot unlearning \cite{DBLP:journals/corr/abs-2201-05629, DBLP:conf/nips/MicaelliS19, DBLP:journals/corr/abs-2111-08947}, respectively.

% In addition to the study of unlearning items, a considerable body of work has delved into the exploration of unlearning specific features or classes. By eliminating the target feature, the classifier's output becomes dissociated from the attribute's value. Existing techniques encompass influence functions \cite{DBLP:conf/ndss/WarneckePWR23} and disentangled representation methods \cite{guo2022efficient}.

In addition to the study of unlearning specific images, a considerable body of work has delved into the exploration of unlearning classes.
Class removal presents a more challenging scenario, requiring the obliteration of data associated with one or multiple classes. Tarun et al. \cite{DBLP:journals/corr/abs-2111-08947} utilized data augmentation for class removal, while Baumhauer et al. \cite{DBLP:journals/ml/BaumhauerSZ22} employed a linear filtration operator that systematically redistributes the classification of images from the class to be forgotten among other classes.

% Except for the study on unlearning the items, plenty of works explore how to unlearn a certain feature or class. With the target feature removed, the output of the classifier is not related to the value of the target feature. Existing techniques include influence functions \cite{DBLP:conf/ndss/WarneckePWR23} and disentangled representation \cite{guo2022efficient}. 
% Class removal is a more challenging scenario where data from a single or even multiple classes ought to be forgotten. Tarun et al. \cite{DBLP:journals/corr/abs-2111-08947} utilized data augmentation for class removal while Baumhauer et al. \cite{DBLP:journals/ml/BaumhauerSZ22} utilized a linear filtration operator that proportionally shifts the classification of the images of the class to be forgotten to other classes.

\subsection{GAN and Machine Unlearning}

The research into GAN models within the realm of machine unlearning is still in its initial stages. One popular direction is to involve the GAN model in the context of classification unlearning. For instance, Youngsik et al. employ a GAN model guided by the target classifier, thereby emulating the training distribution \cite{DBLP:journals/corr/abs-2205-15567}. This innovative approach leads to a more confidential few-shot unlearning. Furthermore, Chen et al. make use of the GAN architecture, inspired by the unlearning intuition that the unlearning data should attain the same performance as the third-party data. The generator acts as the unlearning model, and the discriminator distinguishes the posteriors of unlearning data (the output of the generator) from the posterior of third-party data (the output of the raw generator). When the discriminator cannot distinguish these posteriors, the generator has the same output for unlearning data as the original output for the third-party data, therefore, being the unlearnt version of the original model
\cite{DBLP:journals/corr/abs-2111-11869}.  

Except for the auxiliary role in the classification unlearning context, only one study for GANs focuses on feature unlearning. 
% In a nutshell, Saemi et al. build a map between the original latent code and images which have excluded the target feature \cite{DBLP:journals/corr/abs-2303-05699}. Notably, they construct an implicit feedback system, in which the user can select images containing the target feature among the generated images. In this way, the latent representation of the target feature is identified without access to the training set. 
Saemi et al. \cite{DBLP:journals/corr/abs-2303-05699} try to unlearn a specific feature, such as hairstyle from facial images, from a pre-trained GAN model. After unlearning, the GAN model can not generate images with the unlearned feature or edit the image to exhibit that particular feature. However, when considering item and class unlearning, which are more closely related to practical applications such as personal information withdrawal and addressing inference attacks, there is a notable absence of relevant research.

\section{Preliminaries}

\subsection{Notation}

GAN unlearning indicates that a pre-trained GAN model, comprising a generator $G_0$ and a discriminator $D_0$, selectively forgets some images or an entire class. Notably, these unlearning images are previously used to train the GAN model, the dataset of which is denoted as $D_u$. In contrast, the remaining training images, which are not subjected to unlearning, are referred to as learning images, labeled as $D_l$. The test images are denoted as $D_t$, independent and identically distributed (IID) from the training images.

Within the cascaded unlearning, we propose the substitute mechanism $S(\cdot)$ and fake label $F_{label}$ to facilitate generator and discriminator unlearning. After unlearning, we establish a set of evaluation metrics: $AUC_{l,u}$ and $FID_{u}$ for unlearning effectiveness, $FID_{l}$ for model's intrinsic performance, $ACC$ for the downstream task, and $T$ for the unlearning efficiency. 

\begin{table}[htbp]
  \centering
  \caption{Notations}
    \begin{tabular}{ll}
    \hline
    Notations & Descriptions \\
    \hline
    $G_0$   & Generator of the target GAN \\
    $D_0$   & Discriminator of the target GAN \\
    $G$    & Generator of the unlearning GAN \\
    $D$    & Discriminator of the unlearning GAN \\
    $z$    & Latent code \\
    $D_u$    & The set of images to be forgotten \\
    $D_l$    & The set of images not to be forgotten \\
    $D_t$    & The set of test images \\
    $S(\cdot)$ & The substitute mechanism \\
    $F_{label}$ & \shortstack{The fake label that the discriminator ought\\ to output for the unlearning images}\\
    $AUC_{l,u}$ & The AUC value between $D(x)_{x~D_l}$ and $D(x)_{x~D_u}$ \\
    $FID_{u}$ &The FID distance between the model $G$ and $D_u$ \\
    $FID_{l}$ &The FID distance between the model $G$ and $D_l$ \\
    $ACC$ &The classification accuracy of the downstream task\\
    $T$  & The time the unlearning costs\\
    \hline
    \end{tabular}%
  \label{tab:addlabel}%
\end{table}%

\subsection{Machine Unlearning}

Machine unlearning is the reverse process of learning that a pre-trained model $M$ forgets one or several data records with the unlearning function $U$. The data to be forgotten is denoted as unlearning images $D_u$, which is part of the training set $D_{all}$ as $D_u \in D_{all}$. By contrast, the data not to be forgotten is denoted as learning images $D_l$ that $D_{all} = D_l \cup D_u $. If $M$ successfully unlearns $D_u$, it should not have the same performance when facing the unlearning set $D_u$ and the learning set $D_l$. Suppose we can use function $Q$ to quantize the model performance, a big value denotes better performance. For example, $Q$ could denote the classification accuracy if $M$ as
the classifier or the detection rate if model $M$ for the object detection task. For a GAN model, $Q$ could be the discriminator whose output denotes how confident the GAN model is on the fidelity of the input image. The relationship can be described as:

\begin{equation}
    \begin{aligned}
         Q(M,x)_{x \in D_u} \sim Q(M,x)_{x \in D_l} \\
        Q(U(M),x)_{x \in D_u} < Q(U(M),x)_{x \in D_l} \\
    \end{aligned}
    \label{GAN_hinge_loss}
\end{equation}

There are two classic unlearning methods, including retraining-based machine unlearning \cite{DBLP:conf/sp/BourtouleCCJTZL21, DBLP:conf/alt/Neel0S21} and summation-based machine unlearning \cite{DBLP:conf/sp/CaoY15}. The retraining-based machine unlearning innovates on the training set, aiming for fewer training burdens. For example, the well-known SISA framework \cite{DBLP:conf/sp/BourtouleCCJTZL21}, which splits $D$ into $s$ shards $(D_i)_1^s$ and trains an intermediate model $M_i$ over each shard $D_i$. The final model $M$ is an aggregation of multiple models over these shards. When unlearning images, for example, of $D_1$, it trains the intermediate model $M_1$ alone, which costs significantly fewer resources and time.  

According to \cite{9844865learntoforget}, summation-based machine unlearning focuses on the learning process, i.e., learning through multiple iterations $G_i$, denoted as $M=Learn (G_1,G_2,\dots,G_k)$. By contrast, with images in $D_u$ to be forgotten, the unlearning process is depicted as $M^{'}=Learn (G_1-G_1^{'},G_2-G_2^{'},\dots,G_k-G_k^{'})$. $M^{'}$ is the unlearning version of $M$ which has no memorization of $D_u$. 

\subsection{Generative adversarial networks}

Generative adversarial networks (GANs) simulate the distribution of the training data, thereby, generating plausible images. In detail, a GAN model consists of a generator ($G$) and a discriminator ($D$). $G$ plays the generative role that maps a low-dimensional latent space (such as the latent $Z$ space $p_Z$) to the high-dimensional image space of $x_{'}$ via $x_{'}=G(z)_{z\sim p_Z}$. $D$ plays the regulatory role that distinguishes between the generated image $x_{'}=G(z)_{z\sim p_Z}$ and the training image $x$, as a binary classifier. The output value of $D$ indicates how confident is $G$ on the image fidelity, therefore, higher values for the training images owing to $L_{D_{real}}$ and lower values for the generated images owing to $L_{D_{fake}}$. Formally,

\begin{equation}
    \begin{aligned}
        L_{D_{real}} = & E_{x\sim p_{data}}[log D(x)],\\
        L_{D_{fake}} = & E_{z\sim p_Z}[log(1-D(G(z)))],
    \end{aligned}
\end{equation}
Meanwhile, $G$ tries to deceive $D$ into believing that the generated images are real as the training images, in contrast to $L_{D_{fake}}$. $G$ and $D$ are in a game with the following min-max loss:
\begin{equation}
    \begin{aligned}
        \min L_{G} = & L_{D_{fake}},\\
        \max L_{D} = & L_{D_{real}} + L_{D_{fake}},
    \end{aligned}
    \label{GAN_loss}
\end{equation}

Once training is completed, and given enough capacity, $G$ and $D$ are optimized to their fullest potential. At this point, $D$ reaches its optimum given $G$ at each update, and $G$ is updated to improve the min-max loss. Hence, $G$ can accurately simulate the training data distribution.

% However, the regular GAN loss function may lead to the vanishing gradients problem during the learning process. Least Squares Generative Adversarial Networks (LSGANs) proposed by \cite{DBLP:conf/iccv/MaoLXLWS17} adopt the least squares loss function for the discriminator. Then the objective functions can be defined as follows:
% \begin{equation}
%     \begin{aligned}
%         \max L_{D}^{LSGAN} = & \frac{1}{2} E_{x\sim p_{data}}[(D(x)-b)^2] \\
%         &+ \frac{1}{2} E_{z\sim p_Z}[(D(G(z))-a)^2],\\
%         \min L_{G}^{LSGAN} = & \frac{1}{2} E_{z\sim p_Z}[(D(G(z))-c)^2],\\
%     \end{aligned}
%     \label{GAN_least_loss}
% \end{equation}

% where $a$ and $b$ are the labels for fake data and real data, respectively, and $c$ denotes the value that the generator $G$ wants the discriminator $D$ to believe for fake data. Lim et al. \cite{DBLP:journals/corr/LimY17} combines the hinge loss and regular GAN loss as:

% \begin{equation}
%     \begin{aligned}
%         \max L_{D}^{hinge} = & E_{x\sim p_{data}}[\max(0,1-D(x))] \\
%         &+ E_{z\sim p_Z}[\max(0,1+D(G(z)))],\\
%         \min L_{G}^{hinge} = & E_{z\sim p_Z}[D(G(z))],\\
%     \end{aligned}
%     \label{GAN_hinge_loss}
% \end{equation}
% The hinge loss for the real images penalizes the discriminator when it classifies a real image as fake, and the hinge loss for the fake images penalizes the discriminator when it classifies a fake image as real.

Recently, latent $Z$ space has been accused of entanglement because it must follow the probability density of the training data \cite{DBLP:conf/cvpr/KarrasLA19}. Hence, there's a growing demand for a more disentangled latent space, with the latent $W$ space emerging as a prominent solution. The latent $W$ space is a transformation of the latent $Z$ space through a non-linear mapping network, and it is prominently utilized in StyleGAN \cite{DBLP:conf/cvpr/KarrasLA19} and its related iterations \cite{DBLP:conf/nips/KarrasAHLLA20,DBLP:conf/cvpr/KarrasLAHLA20}. Within these StyleGAN models, the generator consists of two core components: a mapping network $M_{map}$ and a synthesis network $M_{syn}$. The generation process for an image follows the expression $M_{syn}(M_{map}(z))_{z\sim p_Z}$, where $z$ is drawn from the distribution $p_Z$.

\subsection{Membership inference attack}
In a membership inference attack, the adversary deduces whether an image has been utilized to train the target model. Several membership inference algorithms are designed based on the overfitting theory that the trained model has extremely excellent performances on the training image versus test images. For example, the classifier makes decisions with higher confidence scores for training images \cite{Shokri7958568}. With respect to a GAN model, its discriminator and generator have different overfitting performances. The discriminator discriminates training images real with high confidence scores and discriminates test images real with lower confidence scores \cite{hayesLOGANMembershipInference2019}. 
The generator overfits the reconstruction capability in that the reconstruction error of the training images is lower than one of the test images \cite{chenGANLeaksTaxonomyMembership2019,hilprechtMonteCarloReconstruction2019,liuPerformingComembershipAttacks2019}. 

%Existing membership inference attacks on GANs are designed around these overfitting performances, including discriminator-based MIAs such as LOGAN \cite{hayesLOGANMembershipInference2019} and generator-based MIAs such as GANLeaks \cite{chenGANLeaksTaxonomyMembership2019, hilprechtMonteCarloReconstruction2019}. By comparison, the discriminator-based MIA, i.e., LOGAN \cite{hayesLOGANMembershipInference2019} is deemed the most effective because the discriminator has direct access to the training images. 
% Meanwhile, overfitting on GANs raises a controversy that GAN models with evolved architecture such as intermediate latent space suffer little overfitting \cite{websterDetectingOverfittingDeep2019}. 

In this paper, we utilize the membership inference attacks on GANs as an evaluation tool to check if an image has been unlearned. For an unlearning image, if the membership inference attack deduces it in the training set with high possibilities while deducing it in the images with significantly low possibilities after unlearning, the unlearning is successful, according to \cite{9844865learntoforget}. 

\subsection{GAN inversion}

GAN inversion tries to reverse the generation process in which a given image is inverted into the latent space of a pre-trained GAN model and afterward mapped back into image space, denoted as $x = G(\widetilde{G}^{-1}(x))$. $x$ is the original image, $G$ stands for the pre-trained generator, and $\widetilde{G}^{-1}$ stands for the approximate projection operation. \cite{xiaGANInversionSurvey2021} cites three primary types of inversion techniques, including encoder-based inversion which constructs an encoder to discover the mapping from the image space to the latent space \cite{PerarnauWRA16InvertibleConditionalGANsforimageediting,ZhuSZZ20InDomainGAN}; optimization-based inversion which constructs a reconstruction loss function and to optimize over the latent vector \cite{ASpectralRegularizerforUnsupervisedDisentanglement,VoynovB20UnsupervisedDiscoveryofInterpretableDirectionsintheGANLatentSpace}; and hybrid inversion which combines both learning-based and optimization-based methods to use their benefits fully. For instance, Zhu et al. first trained a separate encoder to produce the initial latent code before beginning optimization \cite{ZhuKSE16GenerativeVisualManipulationontheNaturalImageManifold}.

% In a GAN model, there is a weak connection between the training set and the generator. Because the generator has no access to the training set and learns under the guideline of the discriminator. To finish the generator unlearning, we must establish the connection between the specific image and the generator. GAN inversion serves as a dependable technique for achieving this connection. 

% Here, we utilize the target generator $G_0$ and GAN inversion techniques, mapping the unlearning images into the target latent space. Intuitively, it is hard to attain the ground-truth latent code. Suppose we attain $z_0$ through the GAN inversion technique, while $z_0+\beta$ is the ground-truth latent code, $G_0(z_0+\beta) = x_0$ where $\beta$ is subtle. 
% Since the latent space is continuous so if the generator re-builds a mapping $z_0 \rightarrow {x_t}$, the mapping $z_0+\beta \rightarrow x_0$ breakdowns if $x_t$ is quite different from $x_0$.

\subsection{Preliminary challenges of GAN unlearning}
\label{sec:challenges of GAN unlearning}

For GAN unlearning, due to the unique architecture of GAN models, we observe two preliminary challenges to tackle:

% \begin{enumerate}
       
%     \item 
\textbf{Challenge of generator unlearning: disruption of latent space continuity.} The continuity of the latent space dictates that two close points in the latent space should not give two completely different images. The completeness of the latent space means that a point imaged from the latent space should give "meaningful" content. 
According to the overfitting theory, the generator tends to generate training images, denoted as $G_0(z_0) = x_0$, where $x_0$ and $z_0$ denote the unlearning images and its ground-truth latent code. Because they can attain high confident scores from the discriminator. When we try to remove the memory of $x_0$, we actually remove the mapping $z_0 \rightarrow {x_0}$. However, given the inherent continuity of the latent space, so $z_0$ should be mapped to a meaningful image. Determining a suitable substitute for the unlearned image $x_0$ assumes a pivotal role within GAN unlearning.
% (To bypass the dilemma, we design a substitution mechanism that $G(z_0) = S(x_0)$.)

\begin{figure}[tbp]
\centering
\includegraphics[scale=0.5]{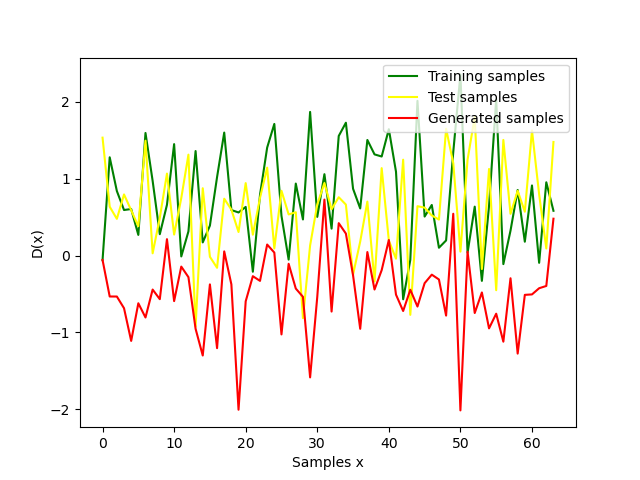}
\caption{The confidence scores that the discriminator of a well-trained GAN model outputs when facing learning images, test images, and generated images}
\label{Fig: mnist_more_D_scores}
\end{figure}

    % \item 
\textbf{The challenge of the discriminator unlearning: how to define a criterion that the discriminator should perform for the unlearning images.} The discriminator $D$ operates as a binary classifier, distinguishing between generated images labeled as fake and training images labeled as real. One might presume that we could draw insights from existing works on classifier unlearning \cite{9844865learntoforget, DBLP:conf/ndss/WarneckePWR23, DBLP:journals/corr/abs-2111-11869}. 
% However, the situation for the discriminator is notably distinct. Firstly, the discriminator faces more complex data. In the unlearning works of classifiers, the target classifier faces the learning and unlearning images. While the discriminator faces the learning images, unlearning images, and generated images. The second difference lies in the criterion. 
In the realm of unlearning for classifiers, the target classifier is expected to exhibit a posterior distribution for unlearning images akin to that of test images (non-member ID images) \cite{9844865learntoforget}. Because the posterior distribution of the training and test images is very different according to the overfit theory \cite{Shokri7958568}. Once the posterior distribution of unlearning images approximates one of the test images, the classifier gets no longer confident in them, thus finishing unlearning. Nevertheless, due to the evolution of GAN models,
the pronounced differences in output between training and test images that discriminators encounter are not as evident in the case of classifiers. Figure \ref{Fig: mnist_more_D_scores} depicts the discriminator's outputs of a StyleGAN model on MNIST dataset. We could observe tangled curves of the training images and test images, and the AUC score of the outputs between the training images and test images is $0.56$. Hence, we cannot utilize the test images for unlearning baseline. Here comes a question: should we employ the output of generated images as the criterion? Remarkably, the AUC score between the outputs of the training images and the generated images is $0.94$.   

% \end{enumerate}

\section{Methodology}

\subsection{Unlearning scenarios}
We introduce few-shot and zero-shot cases of our cascaded unlearning according to the involvement of learning images. In few-shot cascaded unlearning, we have access to three sources of information:
\begin{enumerate}
    \item unlearning images $D_{u}$;
    \item partial learning images $D_l$;
    \item the raw GAN model, including the generator $G_0$ and the discriminator $D_0$.
\end{enumerate}
The learning images provide information that should not be unlearned, which rectifies the over-unlearning. In the zero-shot cascaded unlearning, we have access to the following sources of information:
\begin{enumerate}
    \item unlearning images $D_{un}$;
    \item the raw GAN model, including the generator $G_0$ and the discriminator $D_0$.
\end{enumerate}
In the absence of learning images, zero-shot cascaded unlearning must explore alternative strategies to mitigate the risk of over-unlearning.

Furthermore, we delve into item unlearning and class unlearning, which are the mainstream of unlearning research. In an item unlearning, these unlearning images belong to multiple labels. By contrast, in a class unlearning, these unlearning images belong to a single label, including all images of the unlearning label.

% \begin{table*}[htbp]
%   \centering
%   \caption{Auxiliary information}
%     \begin{tabular}{lrrrrr}
%     \hline
%     \multirow{2}[0]{*}{} & \multicolumn{2}{c}{Training set} & \multicolumn{1}{c}{\multirow{2}[0]{*}{test set}} & \multicolumn{1}{c}{\multirow{2}[0]{*}{Generator}} & \multicolumn{1}{c}{\multirow{2}[0]{*}{Discriminator}} \\
%           & \multicolumn{1}{l}{\shortstack{unlearning set}} & \multicolumn{1}{l}{rest set} &       &       &  \\
%     \hline
%     \shortstack{Unlearn with training set} & 1     & 1     &       & 1     & 1 \\
%     \shortstack{Unlearn without training set} & 1     &       &       & 1     & 1 \\
%     evaluation & 1     & 1     & 1     & 1     & 1 \\
%     \hline
%     \end{tabular}%
%   \label{tab:addlabel}%
% \end{table*}%

\subsection{Unlearning Goals}
\label{sec:Unlearning Goals}

GAN unlearning denotes erasing a pre-trained GAN model's memory of the unlearning dataset. What's more, the GAN model must retain its functionality after unlearning, unless the unlearning is meaningless. Accordingly, we can not conclude a successful unlearning algorithm unless we have evaluated it from the following aspects:

\textbf{Unlearning effectiveness}. A GAN model derives its generative capabilities from the training images. Conversely, when specific images undergo unlearning, the GAN model should eliminate the acquired knowledge derived from these images. In essence, this entails reducing the occurrence of image generation based on unlearned images.

    \begin{itemize}[leftmargin=*]
        \item item unlearning. In item unlearning, it is hard to directly detect whether the target GAN model has successfully unlearned these images solely based on generated images. Because the model retains knowledge from images that adhere to a distribution similar to the unlearned ones. Hence, we evaluate the unlearning effectiveness by observing the discriminator's performance on the unlearning images. 
        \item class unlearning. In class unlearning, we can assess unlearning effectiveness by observing the GAN model's ability to generate high-quality images associated with the unlearned class. For example in a GAN model on MNIST, if the GAN model has unlearnt the class labeled as $7$, it cannot generate meaningful and diverse images of the label $7$.
    \end{itemize}
    % \item \textbf{unlearning effect--overfitting}. According to the overfitting theory, the model performs extremely better on the training images than on the test images. This has been a powerful tool for adversaries to distinguish the training images from test images, namely membership inference attacks. Since we try to erase the memory of the unlearning images, the GAN model should not still overfit on them. In other words, the GAN model performs the same on the unlearning images as on the test images.
    % \item \textbf{unlearning effect--model capability}. A GAN model learns generation capability from the training set. Therefore, if some images are unlearnt, the GAN model has no capability that is used to learn from the unlearning images. This situation is significant in the scenario of class unlearning. For example in a GAN model on MNIST, if the GAN model has unlearnt the number of $7$, it cannot generate meaningful and diverse $7$s.
\textbf{model's intrinsic performance}. After unlearning, the GAN model should perform well based on the capability learned from the remaining training images. By contrast, over-unlearning is a nightmare that makes unlearning and the GAN model meaningless.

\textbf{Downstream tasks' performance}. As we know, GAN models generally play an auxiliary role in various machine learning tasks. For example, a GAN model generates images to augment training data, thereby facilitating a more robust classifier \cite{FRIDADAR2018321}. Hence, the generated images by the unlearned GAN model should continue to effectively serve downstream tasks. This also evaluates the GAN model's performance, however, from the aspect of downstream tasks. 

\textbf{Unlearning efficiency}. The unlearning consumes a shorter time, indicating better efficiency.

\subsection{Substitute mechanism: solution of the first challenge}
\label{sec:substitute mechanism}

Directly deleting the unlearning image $x_0$ would disrupt the continuity of latent space. In this section, we introduce a substitute mechanism $S(\cdot)$, which provides an alternative $S(x_0)$ to the unlearning image, further facilitating an alternative latent-image mapping. In this way, the latent code that originally points to the unlearning images is assigned another meaningful mapping. We follow different rules for the scenarios of item and class unlearning.

Before delving into the details of the substitute rules, it is necessary to determine the ground-truth latent code, denoted as $z_0$, for each unlearning image $x_0$. To achieve this, we employ GAN inversion techniques, which map the unlearning images into the target latent space. This mapping can be expressed as $z_0 = G^{-}(x_0)$, where $G^{-}$ represents the inversion algorithm. While attaining the true ground-truth latent code presents challenges. Let's consider a scenario where we obtain $z_0$ using the GAN inversion technique. In this context, let $z_0 + \beta$ be the actual ground-truth latent code, such that $G_0(z_0+\beta) = x_0$, where $\beta$ is a subtle adjustment. Given the continuity of the latent space, if the generator constructs a mapping $z_0 \rightarrow {S(x_0)}$, the corresponding mapping $z_0+\beta \rightarrow x_0$ breaks down if $S(x_0)$ significantly deviates from $x_0$.

\paragraph{Item unlearning}
In the scenario where single images are to be forgotten, images possibly have different labels. The objective is to perform a form of 'patching' that several mappings change among each class. Here, the substitute $G(\hat{z}$ should meet the requirement that the corresponding latent codes, i.e., $z_0$ and $\hat{z}$ are close, otherwise constructing the new mapping highly destroys other original mappings. We provide three mechanisms, including the average mechanism, the truncation mechanism, and the projection mechanism.

In the \textbf{average mechanism}, we adopt the average image as the substitute, denoted by $\hat{z} = \overline{z}$. In the case of FFHQ, this point represents a sort of average face \cite{DBLP:conf/cvpr/KarrasLA19}. 

The \textbf{truncation mechanism} is a flexible extension of the average mechanism, defined by $\hat{z} = \lambda*z_0+ (1-\lambda) * \overline{z}$. $\lambda$ controls how close $\hat{z}$ is to $z_0$, decreasing from $1$ to $0$. A larger value of $\lambda$ denotes the closer distance. When $\lambda$ is zero, it is the average mechanism. 

The \textbf{projection mechanism} introduces a substitute that evolves gradually, defined as $\hat{z} = \alpha*(z- proj_{\overline{z}(z)}*\overline{z}) +(1-\alpha)*\overline{z}$. Throughout the unlearning process, $\alpha$ diminishes, leading to a fading influence of the attributes associated with the unlearned images. Eventually, $\hat{z}$ converges towards $\overline{z}$. By incorporating the projection mechanism, the model is capable of achieving a gradual unlearning process.
% , which may contribute to enhanced stability.

\paragraph{Class unlearning}
In class unlearning, all images share the same label, denoted as $y_0$. The objective here is a comprehensive update, aimed at preventing the GAN model from generating high-quality images belonging to class $y_0$. We provide two mechanisms, including the average mechanism and the other-class mechanism.

The \textbf{average} mechanism here is the same as the one in the scenario of item unlearning, wherein $\hat{z} = \overline{z}_{y_0}$. Consequently, following the unlearning process, the GAN model becomes capable of generating solely the average image for class $y_0$, while retaining its ability to produce varied images for the remaining classes.

The \textbf{other-class} mechanism advocates for the reorientation of the original latent code $z_0$ towards a different class. To illustrate, consider a set of $10$ classes, and suppose that we intend to forget a class labeled as $7$. In this scenario, for every image $x_0$ belonging to class $7$, the other-class mechanism identifies a class whose average latent code bears the closest resemblance to the corresponding latent code $z_0$ among all other classes. In essence, the average latent code of the identified class becomes the substitute, and it is denoted as follows:

\begin{equation}
    \begin{aligned}
        y_{substitute} =& \arg\min_{y\in Y \setminus y_0} d(z_0,\overline{z}_y) \\
        \hat{z} =& \overline{z}_{y_{substitute}}
    \end{aligned}
    \label{eq:other class sub}
\end{equation}
where $Y$ denotes all the classes, $z_0$ and $y_0$ denote the latent code and label, respectively, of the image $x_0$ slated for unlearning.

\begin{figure}[htbp]
    \centering
    \subfigure[Average mechanism]{
        \includegraphics[scale=1]{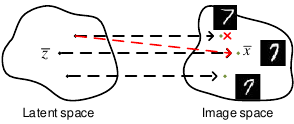}
        \label{fig:average mechanism}}
    \subfigure[Truncation mechanism]{
        \includegraphics[scale=1]{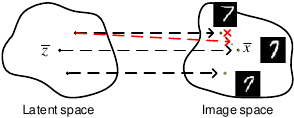}
        \label{fig:truncation mechanism}}
    \subfigure[Projection mechanism]{
        \includegraphics[scale=1]{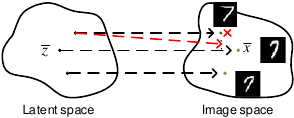}
        \label{fig:projection mechanism}}
    \caption{Substitute mechanism for item unlearning.}
    \label{fig:substitute mechanism for item unlearning}
\end{figure}

\begin{figure}[htbp]
    \centering
    \subfigure[Average mechanism]{
        \includegraphics[scale=1]{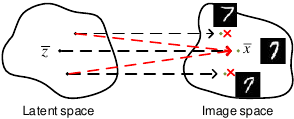}
        \label{fig:average mechanism class}}
    \subfigure[Other class average mechanism]{
        \includegraphics[scale=1]{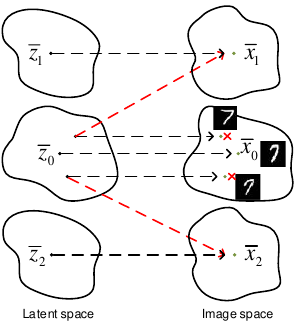}
        \label{fig:other class mechanism}}
    \caption{Substitute mechanism for class unlearning.}
    \label{fig:substitute mechanism for class unlearning}
\end{figure}

\begin{figure}[htbp]
\centering
\includegraphics[scale=0.5]{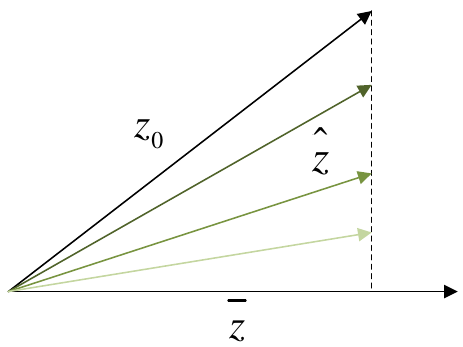}
\caption{Projection mechanism provides a set of gradually changing substitute images.}
\label{Fig:projection}
\end{figure}

\subsection{Fake label: solution of the second challenge}
\label{sec:Fake label}

In section \ref{sec:challenges of GAN unlearning}, we have examined the limitations of employing test images as the reference for completing discriminator unlearning. In this study, we introduce an alternative approach that utilizes a constant value, referred to as the fake label $F_{label}$. We refuse to use the outputs of certain images as the baseline because we observe the significant variance in figure \ref{Fig: mnist_more_D_scores}, which possibly degrades the stability of the algorithm.

To address the specific values of $F_{label}$, we explore several options: [$-1, 0.1, 0.5$]. Then we will explain why we select these values. As discussed in Section \ref{sec:challenges of GAN unlearning}, utilizing the generated images as the baseline emerges as a promising choice. According to the foundational work on GANs \cite{DBLP:conf/nips/GoodfellowPMXWOCB14}, when both the generator and discriminator possess ample capacity, the discriminator becomes incapable of distinguishing between the distributions of the training set and the generated images, yielding $D(x)=0.5$. Hence, we include $0.5$ for $F_{label}$. 
However, achieving this ideal scenario is often challenging. Taking into account real-world scenarios such as in figure \ref{Fig: mnist_more_D_scores}, we include $-1$ for $F_{label}$, representing the minimal value of $D(x)$ of the generated images. Additionally, we incorporate a transitional value, namely $0.1$, to facilitate an intermediate state. When utilized as the baseline for unlearning images, a lower $F_{label}$ signifies reduced $D(x)_{x\sim D_u}$ values, thereby indicating a more intensive unlearning process.

\subsection{Cascaded unlearning algorithm}
\label{sec:Cascaded unlearning algorithm}

We introduce a cascaded unlearning algorithm, comprising both an unlearning algorithm and a learning algorithm. The unlearning algorithm facilitates the GAN model's erasure of specified unlearning images, while the learning algorithm safeguards against over-unlearning. During runtime, the unlearning and learning algorithms operate sequentially. Figure \ref{Fig: cascaded unlearning algorithm} illustrates the structure of the cascaded unlearning algorithm, delineating the unlearning and learning procedures.

\subsubsection{Cascaded unlearning algorithm: few-shot case}
\label{sec:Unlearning with training data}

In the few-shot cascaded unlearning, we have learning images, which enable us to employ the traditional GAN learning algorithm. In this section, we focus on the unlearning algorithm. This unlearning algorithm is inherently designed to steer the GAN model away from generating images akin to those targeted for oblivion. To accomplish this objective, both the discriminator $D$ and the generator $G$ need to forget the unlearning data $D_u$. 
In discriminator unlearning, the devised fake label $F_{label}$ denotes the desired output of the unlearning images from the discriminator $D$.
To attain the desired $F_{label}$ outcome, we leverage the least squared loss $(D(x)-b)^2$ from LSGAN \cite{DBLP:conf/iccv/MaoLXLWS17}. This loss function helps guide the images to positions either away from or around the discriminator's decision boundary, by adjusting the parameter $b$. In a similar vein, with the incorporation of the fake label, we formulate our unlearning loss function for the targeted images as follows:

\begin{equation}
    \begin{aligned}
        L_{D_{unimg}}^{unlearning} = & -E_{x_0\sim p_{un}}[(D(x_0)-F_{label})^2] \\
    \end{aligned}
    \label{eq:GAN_unlearning_loss_D_unimg}
\end{equation}
Meanwhile, the discriminator $D$ also discriminates the generated images as fake, which is achieved by the original loss function $L_{D_{fake}}$. Notably, it could be hinge loss or others that we have no requirements thus making no changes to the original one. Overall, we attain the loss function of the discriminator in unlearning process $L_{D}^{unlearning}$ in the equation \eqref{eq:GAN_unlearning_loss_D}.

% \begin{equation}
%     \begin{aligned}
%         \max L_{D}^{unlearning} = &  \underbrace{-E_{x_0\sim p_{un}}[(D(x_0)-F_{label})^2]}_{L_{D_{unimg}}^{unlearning}} \\
%         & + \underbrace{E_{z\sim p_Z}[log(1-D(G(z)))]}_{L_{D_{fake}}},\\     
%     \end{aligned}
%     \label{eq:GAN_unlearning_loss_D}
% \end{equation}

\begin{equation}
    \begin{aligned}
        \max L_{D}^{unlearning} = L_{D_{unimg}}^{unlearning}
        & + L_{D_{fake}},\\     
    \end{aligned}
    \label{eq:GAN_unlearning_loss_D}
\end{equation}

Smaller output values serve a dual purpose: firstly, they steer the generator away from producing the targeted images; and secondly, they help mitigate the risk of membership inference attacks. Because a significantly high output value is a sign of training data (member) in most membership inference studies \cite{Shokri7958568}. The model unlearns these images intrinsically so adversaries cannot detect image traces from model performance via membership inference attacks.

Generator unlearning aims to deter the generation of images based on the unlearning data. To reach this goal, we first establish the connection between the generator and unlearning images through GAN inversion techniques which find the corresponding latent code for each unlearning image, denoted $G(z_0)=x_0$. Then the proposed substitute mechanism $S(\dot)$ works for generator unlearning which provides a substitute for each unlearning image, denoted $S(x_0)$.
In this way, the latent code $z_0$, which originally points to an unlearning image, points to the substitute image $S(x_0)$. We focus on how to construct the substitute mapping $G(z_0) = S(x_0)$. We consider both pixel-level and perceptual-level distance and mean squared error (MSE) is employed to formalize the distance function, formally,
\begin{equation}
    \begin{aligned}
        L^{unlearning}_{G_{substitute}} = &\lambda_1 *(G(z_0) - S(x_0))^2 \\
        + &\lambda_2 *(VGG(G(z_0)) - VGG(S(x_0)))^2
    \end{aligned}
    \label{eq:GAN_unlearning_loss_G_substitute}
\end{equation}
where $\lambda_1$ and $\lambda_2$ are used to enable and balance the order of magnitude of each loss term. Also, the mapped image $G(z_0)$ should be meaningful and able to deceive the discriminator $D$ to attain high output values. Formally,
\begin{equation}
    \begin{aligned}
        L^{unlearning}_{G_{fake}} = E_{z_0\sim p_{Z_0}} [log(1-D(G(z_0)))]
    \end{aligned}
    \label{eq:GAN_unlearning_loss_G_fake}
\end{equation}
The unlearning loss function of the generator combines $L^{unlearning}_{G_{substitute}}$ and $L^{unlearning}_{G_{fake}}$. Formally,
% \begin{equation}
%     \begin{aligned}
%         \min L_{G}^{unlearning} = &\underbrace{\lambda_1 *(G(z_0) - S(x_0))^2}_{L^{unlearning}_{G_{substitute}}} \\
%         +& \underbrace{\lambda_2 *(VGG(G(z_0)) - VGG(S(x_0)))^2}_{L^{unlearning}_{G_{substitute}}}\\
%          +&\underbrace{E_{z_0\sim p_{Z_0}} [log(1-D(G(z_0)))}_{L^{unlearning}_{G_{fake}}}\\
%     \end{aligned}
%     \label{eq:GAN_unlearning_loss_with_data}
% \end{equation}

\begin{equation}
    \begin{aligned}
        \min L_{G}^{unlearning} =L^{unlearning}_{G_{substitute}} +{L^{unlearning}_{G_{fake}}}\\
    \end{aligned}
    \label{eq:GAN_unlearning_loss_with_data}
\end{equation}

Constructing a substitute mapping $z_0\xrightarrow{G} S(x_0)$ compels the generator $G$ to forget its knowledge of the unlearned images. Simultaneously, this approach preserves the continuity of the latent space because the raw latent codes $z_0$ also point to meaningful images.

\subsubsection{Cascaded unlearning algorithm: zero-shot case}

The zero-shot cascaded unlearning requires no learning images, thus, employing the raw generator $G_0$ and discriminator $D_0$ to guide the learning process. 
% In this section, training images, except for unlearning images, are not accessible, which raises the concern of over-unlearning. Here, we mitigate the concern by using the raw generator $G_0$ and discriminator $D_0$ to guide the learning process. 
The unlearning algorithm in the zero-shot case is the same as elaborated in section \ref{sec:Unlearning with training data}. Hence, our focus here pertains to the learning algorithm.

According to existing machine unlearning literature, GAN models generally play the auxiliary role by generating non-training or analogous-training images \cite{DBLP:journals/corr/abs-2111-11869, DBLP:journals/corr/abs-2205-15567}. This is attributed to the GAN model's inherent capability to generate images that closely approximate real images with remarkable fidelity. Hence, we opt to replace the training images with those generated by the raw generator i.e., $G_0(\dot)$. For the raw generated images $G_0(\dot)$, the discriminator $D$ is supposed to discriminate them as real with high output values. Formally,
\begin{equation}
    \begin{aligned}
        \max L^{learning}_{D_{real}} = E_{z\sim p_Z}[log(D(G_0(z)))]
    \end{aligned}
    \label{eq:GAN_learning_loss_D_real}
\end{equation}

Intuitively, this action inherently weakens the discriminative capability of the discriminator due to the fact that the generated images $G_0(\dot)$ may not possess the same level of fidelity as the original training images. As a countermeasure, we introduce the inclusion of the raw discriminator $D_0$ to guide the current discriminator, with the goal of aligning the discriminator's behavior with that of the original discriminator, $D_0$.
For each instance $G_0(\dot)$ generated by the raw generator, it is desirable for the discriminator to produce a similar output to what the raw discriminator $D_0$ would generate. This process can be conceptualized as a form of knowledge distillation. By implementing this strategy, the current discriminator $D$ gains insights from the knowledge of the raw discriminator $D_0$. Formally, this is expressed as follows:
\begin{equation}
    \begin{aligned}
        L^{learning}_{D_{dist}} = E_{z\sim p_Z}[(D(G_0(z)),D_0(G_0(z)))^2]
    \end{aligned}
    \label{eq:GAN_learning_loss_D_distillation}
\end{equation}
Meanwhile, the discriminator also discriminates the generated images $G(z)$ as fake with small output values. Formally,
\begin{equation}
    \begin{aligned}
        \max L^{learning}_{D_{fake}} = E_{z\sim p_{Z}} [log(1-D(G(z)))]
    \end{aligned}
    \label{eq:GAN_learning_loss_D_fake}
\end{equation}
The learning loss function of the generator combines $L^{learning}_{D_{real}}$, $L^{learning}_{D_{dist}}$ and $L^{learning}_{D_{fake}}$. Formally,
% \begin{equation}
%     \begin{aligned}
%         \max L_{D}^{learning} = & \underbrace{E_{z\sim p_Z}[log(D(G_0(z)))]}_{L^{learning}_{D_{real}}} \\
%         &+ \underbrace{E_{z\sim p_Z}[(D(G_0(z)),D_0(G_0(z)))^2]}_{L^{learning}_{D_{dist}}} \\
%         & + \underbrace{E_{z\sim p_Z}[log(1-D(G(z)))]}_{L^{learning}_{D_{fake}}},\\     
%     \end{aligned}
%     \label{eq:GAN_learning_loss_without_data}
% \end{equation}

\begin{equation}
    \begin{aligned}
        \max L_{D}^{learning} = L^{learning}_{D_{real}}
        &+ L^{learning}_{D_{dist}}
        & + L^{learning}_{D_{fake}},\\     
    \end{aligned}
    \label{eq:GAN_learning_loss_without_data_D}
\end{equation}

When the training of the discriminator $D$ ends, the training of the generator $G$ starts. Here, we adopt the conventional training approach where the generator is updated to produce images capable of deceiving the discriminator. Formally,
% \begin{equation}
%     \begin{aligned} 
%         \min L_{G}^{learning} = &\underbrace{E_{z\sim p_Z}[log(1-D(G(z)))]}_{L_{G}},\\
%     \end{aligned}
%     \label{eq:GAN_learning_loss_without_data}
% \end{equation}
\begin{equation}
    \begin{aligned} 
        \min L_{G}^{learning} = L_{G},\\
    \end{aligned}
    \label{eq:GAN_learning_loss_without_data_G}
\end{equation}

\subsubsection{Summary}
The cascaded unlearning on a GAN model succeeds through successful generator unlearning and discriminator unlearning. The generator establishes a substitute latent-image mapping with the help of our proposed substitute mechanism. And the discriminator becomes less confident in the unlearning images according to the fake label $F_{label}$. Once both generator and discriminator unlearning have been completed, the cascaded unlearning procedure successfully achieves GAN unlearning.

The fundamental difference between the few-shot and zero-shot cascaded unlearning is whether learning images are involved. In the few-shot cascaded unlearning, equivalent learning images are employed to prevent over-unlearning. Conversely, in the zero-shot cascaded unlearning, no learning images but the raw generator $G_0$ are utilized to prevent over-unlearning, which is privacy-preserving.

\section{Experiment}

\subsection{ Data and Setups}
% \subsubsection{Dataset} 
\textbf{Dataset:} Our evaluation encompasses three distinct datasets: MNIST, CIFAR-$10$, and FFHQ, each of which presents an escalating level of image complexity. 
The MNIST dataset comprises $70k$ handwritten digits ($32\times32\times1$) ranging from $0$ to $9$. Among these, $60k$ are allocated for training, and the remaining $10k$ are designated for testing purposes. 
CIFAR-$10$ encompasses 60k color images ($32\times32 \times3$), divided into a training subset of $50k$ images and a testing subset of $10k$ images. This dataset encompasses $10$ distinct classes of objects: plane, car, bird, cat, deer, dog, frog, horse, ship, and truck.
The FFHQ dataset, on the other hand, features $70k$ high-quality human faces. These faces exhibit a wide range of characteristics, including age, gender, ethnicity, and facial expression. Images within the FFHQ dataset can reach up to $256\times256\times3$ pixel dimensions.

% \subsubsection{Target GAN}

\textbf{Target GAN:} With respect to the target GAN, we choose StyleGAN$2$ \cite{DBLP:conf/cvpr/KarrasLAHLA20} architecture. In detail, we involve distinct StyleGAN$2$ models on MNIST, CIFAR-$10$, and FFHQ, respectively. The StyleGAN2 model does not have an inherent encoder, so we adopt the optimized-based inversion technique to establish the connection between the generator and specific images.

\subsection{Evaluation Metric}
% GAN unlearning denotes erasing a pre-trained GAN model's memory of the unlearning dataset $D_u$, namely unlearning goal. Meanwhile, the GAN model's utility should not decline much, including fidelity performance and downstream performance. Note that we consider two unlearning scenarios of item and class unlearning. According to the unlearning goals, we construct a comprehensive evaluation framework.

In this part, we display the evaluation metrics according to the unlearning goals: unlearning effectiveness, model's intrinsic performance, and downstream task performance, and unlearning efficiency.

\textbf{unlearning effectiveness}:
If a GAN model has unlearned certain images, it implies that the model does not master the knowledge gained from those images. Specifically, the discriminator no longer assigns high scores to these images when classifying them as real, and as a result, the generator rarely produces images similar to the ones that have been unlearned.

For item unlearning, we utilize the discriminator's output score $D(\cdot)$ and the Area Under the Curve (AUC) score to quantify the unlearning effectiveness. Formally, we have $AUC_{l,u} = P(D(x)_{x\sim D_l} > D(x)_{x\sim D_u})$, where $D_l$ and $D_u$ denote the learning and unlearning set separately, and $P$ denotes the possibility.
If unlearning succeeds, we would observe a substantial increase in $AUC_{l,u}$.   

To quantify the unlearning effectiveness of class unlearning, except $AUC_{l,u}$, we incorporate the Frechet Inception Distance ($FID(\cdot,\cdot)$) \cite{DBLP:conf/nips/HeuselRUNH17}. Formally, we define $FID_{u}=FID(G(\cdot|y_0),D_u)$, where $y_0$ denotes the unlearning class. A high $FID_{u}$ indicates a long distance between the distribution of the GAN model and the unlearning class, and a successful unlearning as well.

\textbf{Model's intrinsic performance}: we adopt the Frechet Inception Distance to quantify the distance between the distributions of the unlearned GAN model and the learning images, denoted as $FID_{l}=FID(G(\cdot),D_l$. The smaller the increase, the smaller influence of unlearning on the model's intrinsic performance.

\textbf{Downstream task}: we choose the classification as the downstream task and observe the accuracy of images generated by the unlearned GAN model, denoted as $ACC$. $ACC$ is supposed not to decrease much, or the GAN model cannot handle the downstream task of classification after unlearning.

\textbf{Unlearning efficiency}: we evaluate the efficiency by recording the time $T$ that the algorithm takes to complete the unlearning. A shorter duration of unlearning reflects a higher level of efficiency.

% We list all evaluation metrics in table \ref{tab:evaluation metric}.

\textbf{Unlearning destination}
To objectively determine the completion of the model's unlearning, we set specific benchmarks that the unlearning metrics must achieve. In our pursuit of profound unlearning, we set a requirement of $AUC_{l, u} > 0.8$ for item unlearning and $FID_{u} > 150$ for class unlearning.

\subsection{Baseline Method}

Here we adopt the exact unlearning as the baseline, i.e., retraining from scratch. The well-known SISA framework cannot directly extend to GAN models. In the SISA framework, the training set has been split into multiple shards, and each shard is used to train an intermediate model. The aggregation of multiple models over these shards is the final model. When unlearning images, therefore, we only need to retrain the intermediate models that are trained on these unlearning images instead of all. The SISA framework definitely decreases expected unlearning time, achieving more advantageous trade-offs between accuracy and time to unlearn. However, when considering GAN models, the situation differs. A SISA framework for GANs involves multiple pairs of generators and discriminators, with the final output generation being an aggregation of the generators' outputs. Yet, GAN generators produce images rather than confidence scores or labels, making the aggregation of these outputs into meaningful images challenging. In conclusion, the SISA framework does not work for GANs, and retraining from scratch is the baseline of our work.

\subsection{Performance evaluation}

\subsubsection{Item unlearning}
\label{sec:Item unlearning}

We conduct comprehensive experiments on the MNIST, CIFAR-$10$, and FFHQ datasets to evaluate our few-shot and zero-shot cascaded unlearning on specific images. Notably, those images are randomly chosen, thus belonging to diverse classes. 
Table \ref{tab:item unlearning of MNIST}, \ref{tab:item unlearning of CIFAR}, and \ref{tab:item unlearning of FFHQ} display the results of unlearning on the MNIST, CIFAR-$10$, and FFHQ datasets.

\paragraph{Unlearning effectiveness}
\label{sec:item unlearning Unlearning effectiveness}
We observe an overall increase in $AUC_{l, u}$ for all datasets and settings. The increasing values of $AUC_{l, u}$ indicate that the discriminator outputs lower values for the unlearning images compared to the learning images after unlearning. In other words, the discriminator is no longer confident in the unlearning images. Thanks to the min-max training scheme, the generator avoids generating images similar to those unlearning images, ultimately leading to the successful completion of unlearning for the target GAN model.

However, less increase for more complex datasets. For MNIST, both few-shot and zero-shot cascaded unlearning reach the unlearning destination, i.e., $AUC_{l, u} > 0.8$. As for CIFAR-$10$, the few-shot cascaded unlearning successfully achieves the predefined threshold of $AUC_{l, u} > 0.8$; whereas the zero-shot cascaded unlearning encounters challenges, particularly when dealing with a substantial number of images, such as the case with unlearning $1024$ images. Under this scenario, the $AUC_{l,u}$ values can almost increase to $0.76$, $0.74$, and $0.77$ respectively for the three substitute mechanisms, which is notably close to our target threshold of $0.8$. When it comes to the most complex dataset, i.e., FFHQ, $AUC_{l,u}$ is up to round $0.6$, which is far away from our unlearning destination $AUC_{l,u}>0.8$. 
In fact, our unlearning destination suggests an intensive unlearning performance, which is possibly beyond empirical usage. This deliberate choice enables us to rigorously assess the performance of our cascaded unlearning methodology under challenging circumstances. 
In other words, a smaller $AUC_{l,u}$ possibly indicates a successful unlearning. Furthermore, all cases have bypassed the membership inference elaborated in section \ref{sec:Inference evaluation}. In conclusion, our cascaded unlearning achieves effective unlearning on MNIST, CIFAR-$10$, and FFHQ datasets.

\paragraph{Model's intrinsic performance}
According to the $FID_{l}$ increase, unlearning degrades the GAN model's intrinsic performance. The increase is less significant when handling simple datasets such as MNIST or in the few-shot case which involves learning images. For example, in the few-shot cascaded unlearning on MNIST, there is only a marginal increase in $FID_{l}$, and it actually experiences a slight decrease. Intuitively, $FID_{l}$ increases more as more images are forgotten. Specifically, in the few-shot cascaded unlearning on CIFAR-$10$, $FID_{l}$ rises from $5.35$ to approximately $7$ when unlearning $64$ images, and from $5.48$ to around $9$ when unlearning $1024$ images. 

However, we also observe several exceptions. In the zero-shot cascaded unlearning on CIFAR-$10$, shown in table \ref{tab:item unlearning of CIFAR}, the increase in $FID_{l}$ is less pronounced as more images are unlearned. When unlearning $64$ images, $FID_{l}$ escalates from $5.35$ to a minimum of $17.62$. In the case of unlearning $1024$ images, $FID_{l}$ rises from $5.48$ to $8.28$. In the zero-shot cascaded unlearning on FFHQ, shown in table \ref{tab:item unlearning of FFHQ}, unlearning $64$ images induces the highest $FID_{l}$. We speculate the exceptions are related to the learning information,
which is provided by the learning images in the few-shot cascaded unlearning or the raw generator $G_0$ and discriminator $D_0$ in the zero-shot cascaded unlearning. In the cascaded unlearning algorithm, the learning information is equivalent to the unlearning information. In other words, unlearning fewer images indicates less involved learning information, which makes the giant model unstable during the unlearning process.

\paragraph{Downstream task}

This evaluation metric is prepared for datasets with label information, i.e., MNIST and CIFAR-$10$ datasets. we investigate the impact of unlearning on the downstream task using a classifier trained on the entire set of training images. Firstly, we note no substantial decrease in the classification accuracy ($ACC$) on MNIST as well as the few-shot case of CIFAR-$10$. For instance, in the few-shot cascaded unlearning of CIFAR-$10$, $ACC$ increases from $0.75$ to $0.77$ with three substitute mechanisms when unlearning $64$ images and decreases from $0.76$ to $0.74$ with the average substitute mechanism when unlearning $1024$ images.

Next, we have an interesting observation that the classification accuracy ($ACC$) in the zero-shot cascaded unlearning is even higher than those in the few-shot cascaded unlearning. Take the case of unlearning $1024$ MNIST images as an example. $ACC$ values are $0.82$, $0.79$, and $0.75$ in the few-shot case, while $ACC$ values are at least $0.83$ in the zero-shot case. Combined with the higher $FID_{l}$ in the zero-shot case, we conclude that the images generated by the unlearned GAN are less fidelity but keep the classification information well, possibly facilitated by the raw discriminator $D_0$. Because the raw discriminator $D_0$ can provide comprehensive and well-focused classification information while the involved learning images are randomly chosen and provide classification information stochastically.

Additionally, $ACC$ suffers a severe decrease when the model's intrinsic performance downgrades obviously. When unlearning $64$ CIFAR-$10$ images in the zero-shot cascaded unlearning, $FID_{l}$ increases from $5.35$ to at most $21.24$; meanwhile, $ACC$ decrease from $0.75$ to at most $0.66$. The substantial increase in $FID_{l}$ and decrease in $ACC$ suggest that the unlearning has damaged the target GAN model badly.
Additionally, it is worth highlighting that the \textit{average} substitute mechanism demonstrates consistent and reliable performance across both few-shot and zero-shot cascaded unlearning scenarios.

\paragraph{Unlearning efficiency}
\label{sec:Unlearning efficiency}
With respect to the unlearning efficiency, as gauged by $T$, we have made several observations.
$1.$ Zero-shot cascaded unlearning spends more time to reach the unlearning destination of $AUC_{l, u}>0.8$. It is possibly attributed to the decline in the model's intrinsic performance. According to the increasing $FID_{l}$, the discriminator's outputs on learning images decrease. Meanwhile, the outputs on unlearning images decrease, hence, the zero-shot needs more time to reach the specified $AUC_{l,u}$. 
$2.$ More images to unlearn, the more time the algorithm consumes, regardless of few-shot or zero-shot cascaded unlearning. Take the unlearning on MNIST as an example. In the few-shot scenario of MNIST, unlearning $256$ ($1024$) images takes more than $5$ ($38$) times longer than unlearning $64$ images. In the zero-shot scenario of MNIST, unlearning $256$ ($1024$) images costs more than $15$ ($73$) times than unlearning $64$ images. There is an exception on FFHQ. The few-shot cascaded unlearning costs more time to unlearn $64$ images than to unlearn $256$ images; however, still less than to unlearn $1024$ images. This is because the model has been sabotaged badly, whose $FID_{l}$ is up to $21.76$.
$3.$ We observe no significant patterns among the three substitute mechanisms—\textit{average}, \textit{projection}, and \textit{truncation}—with respect to time consumption. Though they differ in several cases.

\paragraph{Summary}
%mnist
In conclusion, we recommend few-shot cascaded unlearning if any learning images are accessible. If not, we recommend unlearning a small number of images within the zero-shot cascaded unlearning.

%cifar
In the few-shot cascaded unlearning, three substitute mechanisms induce an equivalent model performance and downstream task i.e., classification. In the zero-shot cascaded unlearning, the truncation substitute seems to have the best performance. When unlearning $64$ and $256$ images, the unlearning with the truncation substitute induces the lowest $FID_{l}$ and the highest $ACC$ on the premise of completing the unlearning. 

In conclusion, for the complex CIFAR dataset, we recommend involving partial training images and any of three substitute mechanisms if the performances of unlearning, the model itself, and the downstream task are in consideration.

\begin{table*}[htbp]
  \centering
  \caption{Results of item unlearning on MNIST dataset, respectively unlearning $64$, $256$, and $1024$ images. $F_{label}$=0.1, $\lambda_1=1$, and $\lambda_2=0$}
  \subtable[64]{
    \begin{tabular}{lcccccc}
    \hline
    data  & \multicolumn{3}{c}{Few-shot} & \multicolumn{3}{c}{Zero-shot} \\
    substitute & \multicolumn{1}{c}{average} & \multicolumn{1}{c}{projection} & \multicolumn{1}{c}{truncation} & \multicolumn{1}{c}{average} & \multicolumn{1}{c}{projection} & \multicolumn{1}{c}{truncation} \\
    \hline
    \rowcolor{lightgray}
    pre $AUC_{l,u}$ & 0.51  & 0.51  & 0.51  & 0.51  & 0.51  & 0.51  \\
    $AUC_{l,u}$ & 0.86  & 0.86  & 0.85  & 0.81  & 0.81  & 0.81  \\
    % pre $AUC_{u,t}$ & 0.63  & 0.63  & 0.63  & 0.63  & 0.63  & 0.63  \\
    % $AUC_{u,t}$ & 0.21  & 0.19  & 0.19  & 0.17  & 0.15  & 0.15  \\
    \rowcolor{lightgray}
    pre $FID_{l}$ & 3.30  & 3.30  & 3.34  & 3.29  & 3.28  & 3.30  \\
    $FID_{l}$ & 3.43  & 2.09  & 2.20  & 3.67  & 4.87  & 3.36  \\
    \rowcolor{lightgray}
    pre $ACC$ & 0.83  & 0.85  & 0.83  & 0.84  & 0.83  & 0.85  \\
    $ACC$ & 0.81  & 0.79  & 0.79  & 0.84  & 0.85  & 0.85  \\
    $T$ & 360.84  & 312.51  & 311.34  & 948.18  & 791.68  & 637.65  \\
    \hline
    \end{tabular}}
  \subtable[256]{
  \begin{tabular}{lcccccc}
  \hline
    data  & \multicolumn{3}{c}{Few-shot} & \multicolumn{3}{c}{Zero-shot} \\
    substitute & \multicolumn{1}{c}{average} & \multicolumn{1}{c}{projection} & \multicolumn{1}{c}{truncation} & \multicolumn{1}{c}{average} & \multicolumn{1}{c}{projection} & \multicolumn{1}{c}{truncation} \\
    \hline
    \rowcolor{lightgray}
    pre $AUC_{l,u}$ & 0.47  & 0.47  & 0.47  & 0.47  & 0.47  & 0.47  \\
    $AUC_{l,u}$ & 0.86  & 0.81  & 0.83  & 0.81  & 0.81  & 0.80  \\
    % pre $AUC_{u,t}$ & 0.62  & 0.62  & 0.62  & 0.62  & 0.62  & 0.62  \\
    % $AUC_{u,t}$ & 0.20  & 0.23  & 0.21  & 0.19  & 0.16  & 0.18  \\
    \rowcolor{lightgray}
    pre $FID_{l}$ & 3.26  & 3.28  & 3.31  & 3.28  & 3.29  & 3.33  \\
    $FID_{l}$ & 2.34  & 1.86  & 5.01  & 5.82  & 5.66  & 6.60  \\
    \rowcolor{lightgray}
    pre $ACC$ & 0.85  & 0.84  & 0.85  & 0.85  & 0.82  & 0.84  \\
    $ACC$ & 0.82  & 0.85  & 0.86  & 0.84  & 0.83  & 0.88  \\
    $T$ & 2026.99  & 1368.26  & 1403.55  & 14491.18  & 14588.26  & 14167.06  \\
    \hline
    \end{tabular}%
  }
  \subtable[1024]{
    \begin{tabular}{lcccccc}
    \hline
    data  & \multicolumn{3}{c}{Few-shot} & \multicolumn{3}{c}{Zero-shot} \\
    substitute & \multicolumn{1}{c}{average} & \multicolumn{1}{c}{projection} & \multicolumn{1}{c}{truncation} & \multicolumn{1}{c}{average} & \multicolumn{1}{c}{projection} & \multicolumn{1}{c}{truncation} \\
    \hline
    \rowcolor{lightgray}
    pre $AUC_{l,u}$ & 0.50  & 0.50  & 0.50  & 0.50  & 0.50  & 0.50  \\
    $AUC_{l,u}$ & 0.83  & 0.82  & 0.81  & 0.82  & 0.82  & 0.80  \\
    % pre $AUC_{u,t}$ & 0.58  & 0.58  & 0.58  & 0.58  & 0.58  & 0.58  \\
    % $AUC_{u,t}$ & 0.22  & 0.22  & 0.22  & 0.20  & 0.15  & 0.20  \\
    \rowcolor{lightgray}
    pre $FID_{l}$ & 3.31  & 3.28  & 3.29  & 3.28  & 3.32  & 3.30  \\
    $FID_{l}$ & 4.64  & 5.29  & 2.66  & 7.51  & 6.07  & 7.64  \\
    \rowcolor{lightgray}
    pre $ACC$ & 0.81  & 0.84  & 0.82  & 0.83  & 0.84  & 0.84  \\
    $ACC$ & 0.82  & 0.79  & 0.75  & 0.86  & 0.86  & 0.83  \\
    $T$ & 13689.33  & 13269.15  & 13350.87  & 70054.37  & 66166.43  & 58558.20  \\
    \hline
    \end{tabular}}
  \label{tab:item unlearning of MNIST}%
\end{table*}%

% \paragraph{CIFAR-$10$}

% Table generated by Excel2LaTeX from sheet 'Sheet5'
\begin{table*}[htbp]
  \centering
  \caption{Results of item unlearning on CIFAR-$10$ dataset, respectively unlearning $64$, $256$, and $1024$ images. $F_{label}$=0.1, $\lambda_1=1$, and $\lambda_2=0$}
  \subtable[64]{
    \begin{tabular}{lcccccc}
    \hline
    data  & \multicolumn{3}{c}{Few-shot} & \multicolumn{3}{c}{Zero-shot} \\
    substitute & \multicolumn{1}{c}{average} & \multicolumn{1}{c}{projection} & \multicolumn{1}{c}{truncation} & \multicolumn{1}{c}{average} & \multicolumn{1}{c}{projection} & \multicolumn{1}{c}{truncation} \\
    \hline
    \rowcolor{lightgray}
    pre $AUC_{l,u}$ & 0.49  & 0.49  & 0.49  & 0.49  & 0.49  & 0.49  \\
    $AUC_{l,u}$ & 0.99  & 0.99  & 0.98  & 0.82  & 0.83  & 0.80  \\
    % pre $AUC_{u,t}$ & 0.60  & 0.60  & 0.60  & 0.60  & 0.60  & 0.60  \\
    % $AUC_{u,t}$ & 0.17  & 0.18  & 0.17  & 0.20  & 0.20  & 0.20  \\
    \rowcolor{lightgray}
    pre $FID_{l}$ & 5.35  & 5.35  & 5.35  & 5.35  & 5.35  & 5.35  \\
    $FID_{l}$ & 7.84  & 7.28  & 8.28  & 18.59  & 21.24  & 17.62  \\
    \rowcolor{lightgray}
    pre $ACC$ & 0.75  & 0.75  & 0.75  & 0.75  & 0.75  & 0.75  \\
    $ACC$ & 0.77  & 0.77  & 0.77  & 0.66  & 0.66  & 0.65  \\
    $T$ & 502.56  & 494.46  & 492.94  & 741.03  & 760.20  & 767.48  \\
    \hline
    \end{tabular}}
  \subtable[256]{
  \begin{tabular}{lcccccc}
  \hline
    data  & \multicolumn{3}{c}{Few-shot} & \multicolumn{3}{c}{Zero-shot} \\
    substitute & \multicolumn{1}{c}{average} & \multicolumn{1}{c}{projection} & \multicolumn{1}{c}{truncation} & \multicolumn{1}{c}{average} & \multicolumn{1}{c}{projection} & \multicolumn{1}{c}{truncation} \\
    \hline
    \rowcolor{lightgray}
    pre $AUC_{l,u}$ & 0.44  & 0.44  & 0.44  & 0.44  & 0.44  & 0.44  \\
    $AUC_{l,u}$ & 0.99  & 0.97  & 0.97  & 0.79  & 0.80  & 0.80  \\
    % pre $AUC_{u,t}$ & 0.62  & 0.62  & 0.62  & 0.62  & 0.62  & 0.62  \\
    % $AUC_{u,t}$ & 0.55  & 0.52  & 0.53  & 0.16  & 0.16  & 0.17  \\
    \rowcolor{lightgray}
    pre $FID_{l}$ & 5.52  & 5.52  & 5.52  & 5.52  & 5.52  & 5.52  \\
    $FID_{l}$ & 8.98  & 9.36  & 8.75  & 19.97  & 14.87  & 11.49  \\
    \rowcolor{lightgray}
    pre $ACC$ & 0.77  & 0.77  & 0.77  & 0.77  & 0.77  & 0.77  \\
    $ACC$ & 0.77  & 0.78  & 0.77  & 0.69  & 0.73  & 0.77  \\
    $T$ & 1530.59  & 1544.15  & 1528.43  & 11553.41  & 6516.46  & 6616.05  \\
    \hline
    \end{tabular}%
  }
  \subtable[1024]{
  \begin{tabular}{lcccccc}
  \hline
    data  & \multicolumn{3}{c}{Few-shot} & \multicolumn{3}{c}{Zero-shot} \\
    substitute & \multicolumn{1}{c}{average} & \multicolumn{1}{c}{projection} & \multicolumn{1}{c}{truncation} & \multicolumn{1}{c}{average} & \multicolumn{1}{c}{projection} & \multicolumn{1}{c}{truncation} \\
    \hline
    \rowcolor{lightgray}
    pre $AUC_{l,u}$ & 0.49  & 0.49  & 0.49  & 0.49  & 0.49  & 0.49  \\
    $AUC_{l,u}$ & 1.00  & 1.00  & 1.00  & 0.76  & 0.74  & 0.77  \\
    % pre $AUC_{u,t}$ & 0.60  & 0.60  & 0.60  & 0.60  & 0.60  & 0.60  \\
    % $AUC_{u,t}$ & 0.54  & 0.54  & 0.46  & 0.18  & 0.21  & 0.17  \\
    \rowcolor{lightgray}
    pre $FID_{l}$ & 5.48  & 5.48  & 5.48  & 5.48  & 5.47  & 5.48  \\
    $FID_{l}$ & 9.51  & 9.55  & 9.48  & 8.28  & 8.56  & 10.11  \\
    \rowcolor{lightgray}
    pre $ACC$ & 0.76  & 0.75  & 0.75  & 0.75  & 0.75  & 0.75  \\
    $ACC$ & 0.74  & 0.73  & 0.75  & 0.80  & 0.74  & 0.79  \\
    $T$ & 12978.90  & 12981.60  & 17216.22   & 60273.76  & 69533.64  & 82248.14  \\
    \hline
    \end{tabular}%
  }
  \label{tab:item unlearning of CIFAR}%
\end{table*}%

% \paragraph{FFHQ}

% Table generated by Excel2LaTeX from sheet 'Sheet (2)'
\begin{table}[htbp]
  \centering
  \caption{Results of item unlearning on FFHQ dataset, respectively unlearning $64$, $256$, and $1024$ images. Using few-shot and \textit{average} substitute mechanism with $F_{label}$=0.1, $\lambda_1=1$, and $\lambda_2=0$}
    \begin{tabular}{lccc}
    \hline
    Unlearn num & \multicolumn{1}{c}{64} & \multicolumn{1}{c}{256} & \multicolumn{1}{c}{1024} \\
    \hline
    \rowcolor{lightgray}
    pre $AUC_{l,u}$ &  0.57  & 0.50  & 0.50 \\
    $AUC_{l,u}$ & 0.61  & 0.59  & 0.61 \\
    \rowcolor{lightgray}
    pre $FID_l$ & 12.42  & 12.43  & 12.42  \\
    $FID_l$ & 21.76  & 13.70  & 16.61 \\
    % \rowcolor{lightgray}
    % pre unlearn-test auc & 0.58  & 0.68  & 0.69  \\
    % unlearn-test auc & 0.42  & 0.55  & 0.50  \\
    $T$ & 300175.07  & 157989.13  & 380806.56  \\
    \hline
    \end{tabular}%
  \label{tab:item unlearning of FFHQ}%
\end{table}%

\subsubsection{class unlearning}

In this section, we evaluate our few-shot and zero-shot cascaded unlearning of class unlearning on the MNIST and CIFAR-$10$ datasets. Class unlearning denotes that all images of the unlearning label should be forgotten. In other words, the GAN model cannot generate images with high fidelity that belong to the unlearning label. For the experiments on MNIST and CIFAR-$10$ datasets, we choose the class labeled $7$ to unlearn.
Table \ref{tab:class unlearning of MNIST} and \ref{tab:class unlearning of CIFAR-$10$} display the results of unlearning on the MNIST and CIFAR-$10$ datasets.

\paragraph{Unlearning effectiveness}

At first, both the few-shot and zero-shot cascaded unlearning has achieved successful unlearning on MNIST and CIFAR-$10$ datasets, according to $FID_{u}$ and $AUC_{l,u}$ values. All $FID_{u}$ values exceed our set destination, i.e., $150$, indicating the generated images of the chosen label are far from fidelity. Meanwhile, $AUC_{l,u}$ values are at least $0.86$, indicating the discriminator outputs for the unlearning class images are much lower than the outputs for the images of the left classes. Therefore, the generator won't generate high-fidelity images of the unlearning class under the guidance of the discriminator.

\begin{figure}[htbp]
    \centering
    \subfigure[Before unlearn]{
        \includegraphics[scale=0.42]{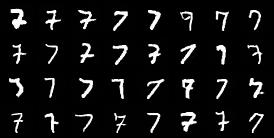}}
    \subfigure[Before unlearn]{
        \includegraphics[scale=0.42]{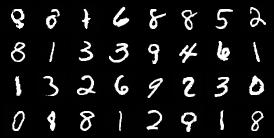}}
    \subfigure[Few-shot (average)]{
    \includegraphics[scale=0.42]{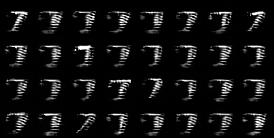}}
    \subfigure[Few-shot (average)]{
    \includegraphics[scale=0.42]{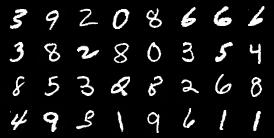}}
    \subfigure[Few-shot (other-class)]{
    \includegraphics[scale=0.42]{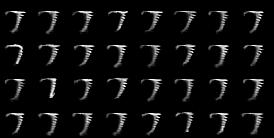}
    }
    \subfigure[Few-shot (other-class)]{
    \includegraphics[scale=0.42]{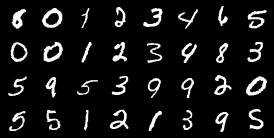}}
    \subfigure[Zero-shot (average)]{
    \includegraphics[scale=0.42]{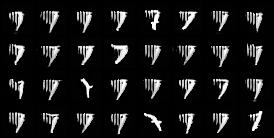}}
    \subfigure[Zero-shot (average)]{
    \includegraphics[scale=0.42]{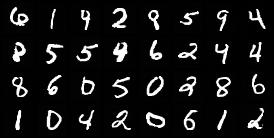}}
    \subfigure[Zero-shot (other-class)]{
    \includegraphics[scale=0.42]{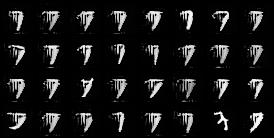}}
    \subfigure[Zero-shot (other-class)]{
    \includegraphics[scale=0.42]{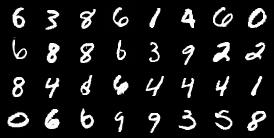}}
    \caption{Visual results of class unlearning on MNIST dataset and the unlearning label is $7$, including generated images of unlearning and other labels before and after few-shot and zero-shot cascaded unlearning. The left column is of unlearning label i.e., $7$ and the right column is of the other labels. And the first row comes from the target GAN without unlearning operations. $F_{label}$=0.1, $\lambda_1=1$, and $\lambda_2=0$.}
    \label{fig: unlearning class of MNIST}
\end{figure}

Figure \ref{fig: unlearning class of MNIST} and \ref{tab:class unlearning of CIFAR-$10$} provide the visual proof, depicting the generated images of the unlearning and left labels. The left column displays the generated images of the unlearning class, and the right column displays the generated images of other classes. The first row displays the generated images of the unlearning and other classes before unlearning for reference. Obviously, in all cases, the generated images of the unlearning class (the left column) are out of fidelity compared with the original ones (the top left) and the images of other classes (the right column). For CIFAR-$10$ dataset within the zero-shot cascaded unlearning, the generated images of the unlearning class are less fidelity though its $AUC_{l,u}$ is lower than the one in the few-shot cascaded unlearning.

\paragraph{Model's intrinsic performance}
As for the model's intrinsic performance, we observe $FID_{l}$ to quantify the generation capability of other classes. A higher $FID_{l}$ denotes a poorer generation capability. For the MNIST dataset in \ref{tab:class unlearning of MNIST}, statistically, the few-shot cascaded unlearning induces equivalent $FID_{l}$ values; while the zero-shot cascaded unlearning induces higher $FID_{l}$ values from round $3$ to above $6$, indicating the model's generation capability definitely degrades. Such degradation has no significant impact on the visual way, by observing the generated images of the other classes in the right column in figure \ref{fig: unlearning class of MNIST}. 
However, for the complex CIFAR-$10$ dataset in \ref{tab:class unlearning of CIFAR-$10$}, the few-shot and zero-shot cascaded unlearning degrade the model's intrinsic performance with obviously increasing $FID_{l}$. And few-shot does not take any advantages from the involvement of learning images, whose $FID_{l}$ is $10.87$ and $13.79$ while $FID_{l}$ of the zero-shot case is $11.26$ and $9.70$ for \textit{average} and \textit{other-class-average} substitute mechanisms.

% MNIST
% As for the model's intrinsic performance, we observe $FID_{l}$ to quantify the generation capability of other classes. A higher $FID_{l}$ denotes a poorer generation capability. Statistically, the few-shot cascaded unlearning induces equivalent $FID_{l}$ values. While the zero-shot cascaded unlearning induces higher $FID_{l}$ values from round $3$ to above $6$, indicating the model's generation capability definitely degrades. Such degradation has no significant impact on the visual way, by observing the generated images of the other classes in the right column in figure \ref{fig: unlearning class of MNIST}. 

% CIFAR

% The few-shot and zero-shot cascaded unlearning degrade the model's intrinsic performance with obviously increasing $FID_{l}$. And few-shot does not take any advantages from the involvement of learning images, whose $FID_{l}$ is $10.87$ and $13.79$ while $FID_{l}$ of the zero-shot case is $11.26$ and $9.70$ for \textit{average} and \textit{other-class-average} substitute mechanisms. 
% Combined with $T$ that few-shot cascaded unlearning spends more than twice times, we conclude that long-time unlearning disrupts the model's intrinsic performance.

\paragraph{Downstream task}
We observe that using our cascaded unlearning algorithm to unlearn a class has a minor influence on downstream tasks such as classification. For MNIST, the few-shot and zero-shot cascaded unlearning even improve the downstream task performance in most cases, reporting increasing $ACC$ values. In the few-shot cascaded unlearning, $ACC$ increases from $0.83$ to $0.85$ with \textit{average} substitute mechanism and from $0.82$ to $0.84$ with \textit{the-other-average} substitute mechanism. In the zero-shot cascaded unlearning, $ACC$ also increases from $0.82$ to $0.86$ with \textit{the-other-average} substitute mechanism and does not decrease much with \textit{average} substitute mechanism, i.e., from $0.85$ to $0.84$.
For CIFAR-$10$, the downstream task performance decreases a bit in the few-shot cascaded unlearning but keeps well in the zero-shot cascaded unlearning. In the few-shot cascaded unlearning, $ACC$ decreases from $0.78$ to $0.75$ with \textit{average} substitute mechanism and from $0.78$ to $0.73$ with \textit{the-other-average} substitute mechanism. In the zero-shot cascaded unlearning, $ACC$ is $0.80$ and $0.78$ for \textit{average} and \textit{the-other-average} substitute mechanisms. Notably, $ACC$ values with class unlearning are higher than when unlearning $1024$ images. Therefore, we conclude that the target GAN model can handle the downstream tasks such as classification, after using our cascaded unlearning in few-shot or zero-shot cases to unlearn a class.

% MNIST
% Interestingly, the few-shot and zero-shot cascaded unlearning even improve the downstream task performance in most cases, reporting increasing $ACC$ values. In the few-shot cascaded unlearning, $ACC$ increases from $0.83$ to $0.85$ with \textit{average} substitute mechanism and from $0.82$ to $0.84$ with \textit{the-other-average} substitute mechanism. In the zero-shot cascaded unlearning, $ACC$ also increases from $0.82$ to $0.86$ with \textit{the-other-average} substitute mechanism which does not decrease much, i.e., from $0.83$ to $0.85$ with \textit{average} substitute mechanism.

% CIFAR
% The downstream task performance decreases a bit in the few-shot cascaded unlearning but keeps well in the zero-shot cascaded unlearning. In the few-shot cascaded unlearning, $ACC$ decreases from $0.78$ to $0.75$ with \textit{average} substitute mechanism and from $0.78$ to $0.73$ with \textit{the-other-average} substitute mechanism. In the zero-shot cascaded unlearning, $ACC$ is $0.80$ and $0.78$ for \textit{average} and \textit{the-other-average} substitute mechanisms.

\paragraph{Unlearning efficiency}
We have different observations of unlearning efficiency from the item unlearning:
$1.$ The few-shot cascaded unlearning needs more time to reach the set destination. It is the opposite in the item unlearning, but reasonable. Firstly, as described in section \ref{sec:Cascaded unlearning algorithm}, the involvement of the training images is to prevent over-unlearning, preventing unlearning as well. Furthermore, the class unlearning utilizes $FID_{u}$ as the destination, instead of $AUC_{l,u}$, which is the attribution that the zero-shot case spends more time in section \ref{sec:Unlearning efficiency}. 
$2.$ The cascaded unlearning with the \textit{other-class} substitute mechanism spends more time. In the few-shot cascaded unlearning of MNIST and CIFAR-$10$ datasets, the \textit{other-class} substitute mechanism even spends almost twice as much time as the \textit{average} substitute mechanism. However, the large time consumption does not always facilitate the unlearning. For example, in the few-shot cascaded unlearning in CIFAR-$10$ in table \ref{tab:class unlearning of CIFAR-$10$}, the \textit{other-class} substitute mechanism spends $1.83$ times as much time as the \textit{average} substitute mechanism, however, having a higher $FID_{l}$ and lower $ACC$ value. The two metric values indicate both a subpar model's intrinsic performance and a detrimental impact on the downstream tasks such as classification.
$3.$ Compared with the item unlearning, class unlearning seems to be more efficient, especially for the zero-shot scenario. Take the MNIST as an example. The class unlearning in the zero-shot case spends comparative or even less time than unlearning $256$ or $1024$ times in the few-shot or zero-shot case.                                                                                                 

% With the zero-shot cascaded unlearning, unlearning a class spends less time than unlearning $256$ images, however, at the cost of the model's intrinsic performance with higher $FID_{l}$s.

% MNIST
% Lastly, we focus on unlearning efficiency. We record the time the unlearning consumes in table \ref{tab:class unlearning of MNIST}. The few-shot cascaded unlearning needs more time to reach the set destination. As described in section \ref{sec:Cascaded unlearning algorithm}, the involvement of the training images is to prevent over-unlearning, preventing unlearning as well. Additionally, in the few-shot cascaded unlearning, \textit{the other-class} substitute mechanism even spends more time, almost twice as much as the \textit{average} substitute mechanism. 

% CIFAR
% Lastly, we focus on unlearning efficiency. we have similar observations on CIFAR-$10$ dataset to ones on MNIST datset: $1.$
% The few-shot cascaded unlearning needs more time to reach the set destination. 
% $2.$ In the few-shot cascaded unlearning, the \textit{other-class-average} substitute mechanism spends almost twice as much as the \textit{average} substitute mechanism. 
% $3.$ The zero-shot cascaded unlearning spends less when unlearning a class than unlearning $1024$ images.

\paragraph{summary}

Both the few-shot and zero-shot cascaded unlearning achieve a successful class unlearning, and \textit{average} substitute mechanism has a stable performance overall. From the perspective of privacy protection, we recommend zero-shot cascaded unlearning, which has advantages in both downstream tasks such as classification and unlearning efficiency.

% For MNIST, both the few-shot and zero-shot cascaded unlearning achieve a successful class unlearning. With respect to the model's intrinsic performance and unlearning efficiency, the few-shot with \textit{average} substitute mechanism has the best performance overall.

% CIFAR
% In conclusion, for a complex dataset such as CIFAR-$10$, we recommend zero-shot with \textit{average} substitute mechanism, which involves no learning images to prevent privacy leakage, ensures model's intrinsic performance and downstream task performance to some extent, and is quite efficient.

% \paragraph{MNIST}

% In conclusion, for the class unlearning of the MNIST dataset, 
% if we pursue the performances of the model itself and the downstream task on the premise of successful unlearning, the involvement of training images and the average substitute mechanism are recommended for efficiency and effectiveness. 

% Table generated by Excel2LaTeX from sheet 'Sheet5'
\begin{table}[htbp]
  \centering
  \caption{Results of class unlearning on MNIST dataset and the unlearning label is $7$. $F_{label}$=0.1, $\lambda_1=1$, and $\lambda_2=0$.}
  \resizebox*{0.5\textwidth}{!}{
    \begin{tabular}{lcccc}
    \hline
    data  & \multicolumn{2}{c}{Few-shot} & \multicolumn{2}{c}{Zero-shot} \\
    substitute & \multicolumn{1}{c}{average} & \multicolumn{1}{c}{other-class} & \multicolumn{1}{c}{average} & \multicolumn{1}{c}{other-class} \\
    \hline
    \rowcolor{lightgray}
    pre $FID_u$ & 6.41  & 6.47  & 6.35  & 6.41  \\
    $FID_u$ & 155.32  & 157.27  & 161.12  & 183.09  \\
    \rowcolor{lightgray}
    pre $AUC_{l,u}$ & 0.21  & 0.21  & 0.21  & 0.21  \\
    $AUC_{l,u}$ & 1.00  & 1.00  & 1.00  & 1.00  \\
    \rowcolor{lightgray}
    pre $FID_{l}$ & 3.51  & 3.51  & 3.46  & 3.44  \\
    $FID_{l}$ & 3.97  & 1.96  & 6.12  & 8.89  \\
    \rowcolor{lightgray}
    pre $ACC$ & 0.83  & 0.82  & 0.85  & 0.82  \\
    $ACC$ & 0.85  & 0.84  & 0.84  & 0.86  \\
    $T$ & 7674.45  & 12612.06  & 1944.07  & 2070.28  \\
    \hline
    \end{tabular}%
    }
  \label{tab:class unlearning of MNIST}%
\end{table}%

% \paragraph{CIFAR-$10$}

\begin{table}[htbp]
  \centering
  \caption{Results of class unlearning on CIFAR-$10$ dataset and the unlearning label is $7$. $F_{label}$=0.1, $\lambda_1=1$, and $\lambda_2=0$.}
  \resizebox*{0.5\textwidth}{!}{
    \begin{tabular}{lcccc}
    \hline
    data  & \multicolumn{2}{c}{Few-shot} & \multicolumn{2}{c}{Zero-shot} \\
    substitute & \multicolumn{1}{c}{average} & \multicolumn{1}{c}{other-class} & \multicolumn{1}{c}{average} & \multicolumn{1}{c}{other-class} \\
    \hline
    % perceptual\_loss\_lamda & \multicolumn{1}{c}{0} & \multicolumn{1}{c}{0} & \multicolumn{1}{c}{0} & \multicolumn{1}{c}{0} \\
    \rowcolor{lightgray}
    pre $FID_{u}$ & 12.47  & 12.47  & 12.47  & 12.47  \\
    $FID_{u}$ & 152.02  & 150.25  & 155.73  & 150.59  \\
    \rowcolor{lightgray}
    pre $AUC_{l,u}$ & 0.62  & 0.62  & 0.62  & 0.62  \\
    $AUC_{l,u}$ & 1.00  & 1.00  & 0.86  & 0.87  \\
    \rowcolor{lightgray}
    pre $FID_{l}$ & 5.60  & 5.60  & 5.60  & 5.60  \\
    $FID_{l}$ & 10.87  & 13.79  & 11.26  & 9.70  \\
    \rowcolor{lightgray}
    pre $ACC$ & 0.78  & 0.78  & 0.78  & 0.78  \\
    $ACC$ & 0.75  & 0.73  & 0.80  & 0.78  \\
    $T$ & 35060.72  & 64304.34  & 14099.00  & 20693.14  \\
    \hline
    \end{tabular}%
    }
  \label{tab:class unlearning of CIFAR-$10$}%
\end{table}%

\begin{figure}[htbp]
    \centering
    \subfigure[Before unlearn]{
        \includegraphics[scale=0.42]{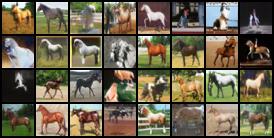}}
    \subfigure[Before unlearn]{
        \includegraphics[scale=0.42]{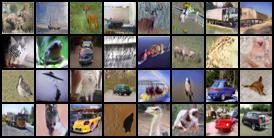}}
    \subfigure[Few-shot (average)]{
    \includegraphics[scale=0.42]{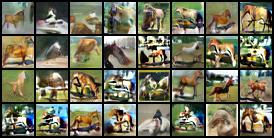}}
    \subfigure[Few-shot (average)]{
    \includegraphics[scale=0.42]{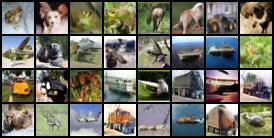}}
    \subfigure[Few-shot (other-class)]{
    \includegraphics[scale=0.42]{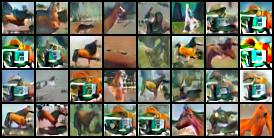}}
    \subfigure[Few-shot (other-class)]{
    \includegraphics[scale=0.42]{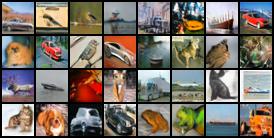}}
    \subfigure[Zero-shot (average)]{
    \includegraphics[scale=0.42]{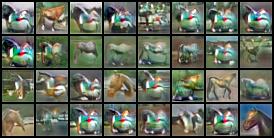}}
    \subfigure[Zero-shot (average)]{
    \includegraphics[scale=0.42]{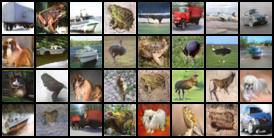}}
    \subfigure[Zero-shot (other-class)]{
    \includegraphics[scale=0.42]{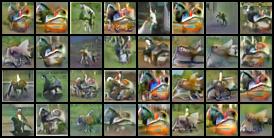}}
    \subfigure[Zero-shot (other-class)]{
    \includegraphics[scale=0.42]{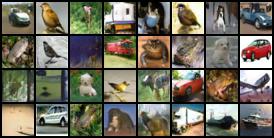}}
    \caption{Visual results of class unlearning on CIFAR-$10$ dataset and the unlearning label is $7$, including generated images of unlearning and other labels before and after few-shot and zero-shot cascaded unlearning. The left column is of unlearning label i.e., $7$ and the right column is of the other labels. And the first row comes from the target GAN without unlearning operations. $F_{label}$=0.1, $\lambda_1=1$, and $\lambda_2=0$.}
    \label{fig: unlearning class of CIFAR-$10$}
\end{figure}

\subsection{Parameter evaluation}

\subsubsection{The criterion of discriminator unlearning: Fake label}

As mentioned in section \ref{sec:Fake label}, fake label $F_{label}$ is the baseline value that the discriminator is expected to output for the unlearning images when unlearning finishes, with possible values of $-1$, $0.1$, and $0.5$. Theoretically, a higher $F_{label}$ suggests a moderate unlearning and a small one suggests an intensive unlearning and vice versa.
In this part, we validate the assumption.

% , which is lower than the general value that the discriminator outputs for the unlearning images before unlearning starts. Hence, a higher $F_{label}$ theoretically suggests a moderate unlearning and a small one suggests an intensive unlearning and vice versa.

Table \ref{tab:item unlearning of mnist: fake label} and \ref{tab:item unlearning of CIFAR-$10$: fake label} separately shows the statistical results of item unlearning on MNIST and CIAFR10 datasets, depicting the average performance of three substitute mechanisms. 
When $F_{label}=-1$, we definitely finish intensive unlearning with the highest $AUC_{l,u}$ and lowest $T$. However, the intensity causes steep degradation in the model's intrinsic performance with increasing $FID_{l}$ and downstream task with decreasing $ACC$. In a nutshell, $F_{label}=-1$ can finish an unlearning but make no sense for the model self or downstream task.
When $F_{label}=0.5$, zero-shot cascaded unlearning fails to finish unlearning on MNIST dataset, attaining an increasing $AUC_{l,u}$ and an extremely high $FID_{l}$. In other cases of $F_{label}=0.5$, the unlearning is successful and even attains the best performance in the model self and downstream task. Nevertheless, we do not recommend this setting because it takes extremely more time and possibly fails in unlearning. 
By comparison, $F_{label}=0.1$ strikes a favorable trade-off between unlearning effectiveness and efficiency, as well as other performance factors such as the model's intrinsic performance and downstream task performance. 

\begin{table}[htbp]
  \centering
  \caption{Statistical results with changing fake labels $F_{label}$ on MNIST. The cascaded unlearning unlearns $64$ images with $F_{label}=[-1,0.1, 0.5]$, $\lambda_1=1$, and $\lambda_2=0$.}
  \resizebox*{0.5\textwidth}{!}{
    \begin{tabular}{lrrrrrr}
    \hline
    & \multicolumn{3}{c}{Few-shot} & \multicolumn{3}{c}{Zero-shot} \\
   $F_{label}$ & \multicolumn{1}{l}{-1} & \multicolumn{1}{l}{0.1} & \multicolumn{1}{l}{0.5} & \multicolumn{1}{l}{-1} & \multicolumn{1}{l}{0.1} & \multicolumn{1}{l}{0.5} \\
    \hline
    \rowcolor{lightgray}
    pre $AUC_{l,u}$ & 0.51  & 0.51  & 0.51  & 0.51  & 0.51  & 0.51  \\
    $AUC_{l,u}$ & 0.93  & 0.86  & 0.83  & 0.86  & 0.81  & 0.32  \\
    \rowcolor{lightgray}
    pre $FID_{l}$ & 3.29  & 3.32  & 3.30  & 3.31  & 3.29  & 3.31  \\
    $FID_{l}$ & 9.81  & 2.57  & 3.77  & 5.99  & 3.97  & 23.17  \\
    \rowcolor{lightgray}
    pre $ACC$ & 0.84  & 0.84  & 0.84  & 0.82  & 0.84  & 0.83  \\
    $ACC$ & 0.65  & 0.80  & 0.83  & 0.82  & 0.85  & 0.76  \\
    $T$ & 322.10  & 328.23  & 1273.63  & 366.60  & 792.50  & 7019.96  \\
    \hline
    \end{tabular}%
    }
  \label{tab:item unlearning of mnist: fake label}%
\end{table}%

% Table generated by Excel2LaTeX from sheet '整理后'
\begin{table}[htbp]
  \centering
  \caption{Statistical results with changing fake labels $F_{label}$ on CIFAR-$10$. The cascaded unlearning unlearns $64$ images with $F_{label}=[-1,0.1, 0.5]$, $\lambda_1=1$, and $\lambda_2=0$.}
  \resizebox*{0.5\textwidth}{!}{
    \begin{tabular}{lrrrrrr}
    \hline 
          & \multicolumn{3}{c}{Few-shot} & \multicolumn{3}{c}{Zero-shot} \\
    $F_{label}$     & \multicolumn{1}{l}{-1} & \multicolumn{1}{l}{0.1} & \multicolumn{1}{l}{0.5} & \multicolumn{1}{l}{-1} & \multicolumn{1}{l}{0.1} & \multicolumn{1}{l}{0.5} \\
    \hline
    \rowcolor{lightgray}
    pre $AUC_{l,u}$ & 0.49  & 0.49  & 0.49  & 0.49  & 0.49  & 0.49  \\
    $AUC_{l,u}$ & 1.00  & 0.98  & 0.92  & 0.93  & 0.82  & 0.82  \\
    \rowcolor{lightgray}
    pre $FID_{l}$ & 5.35  & 5.35  & 5.35  & 5.35  & 5.35  & 5.35  \\
    $FID_{l}$ & 27.16  & 7.80  & 7.14  & 22.77  & 19.15  & 15.49  \\
    \rowcolor{lightgray}
    pre $ACC$ & 0.75  & 0.75  & 0.75  & 0.75  & 0.75  & 0.75  \\
    $ACC$ & 0.67  & 0.77  & 0.77  & 0.62  & 0.66  & 0.74  \\
    $T$ & 573.50  & 496.65  & 591.56  & 527.84  & 756.24  & 5360.60  \\
    \hline
    \end{tabular}%
    }
  \label{tab:item unlearning of CIFAR-$10$: fake label}%
\end{table}%

\subsubsection{Distance metric in the unlearning process: perceptual level distance}

As a crucial component of the cascaded unlearning approach, the unlearning process is designed to remove specific unlearning images or classes from a pre-trained GAN model. It leverages the alternative image provided by the substitute mechanism to establish an alternative latent-image mapping, denoted as $z_0 \rightarrow S(x_0)$, where $x_0$ represents the image to be unlearned and $z_0$ is its corresponding latent code. When creating this alternative mapping, we take into account both pixel-level and perceptual-level distances, which are individually controlled by the parameters $\lambda_1$ and $\lambda_2$. In this section, we maintain the value of $\lambda_1$ at $1$ and vary the value of $\lambda_2$ to explore the influence of the distance metric on the unlearning performance.

The incorporation of perceptual-level distance, in conjunction with pixel-level distance, facilitates a precise and efficient alternative mapping from the original latent code $z_0$, which corresponds to the unlearning image $x_0$, to the image provided by the substitute mechanism $S(\cdot)$. The perceptual-level distance is particularly suited for handling complex images, such as those found in the CIFAR-$10$ dataset. We utilize the parameter $\lambda_2$ to regulate its influence, exploring values of $0$, $0.5$, and $1$. A higher $\lambda_2$ indicates increased involvement of perceptual-level distance, while $\lambda_2=0$ implies no utilization. Theoretically, incorporating perceptual loss enhances efficiency, as a more accurate distance function expedites the unlearning substitute process. To comprehensively investigate the impact of perceptual loss, we perform few-shot cascaded unlearning for both item unlearning and class unlearning on the CIFAR-$10$ dataset.

\paragraph{item unlearning}
The statistical results of item unlearning are presented in table \ref{tab:item unlearning of CIFAR perceptual loss}. With all values of $\lambda_2$, our cascaded unlearning has achieved successful unlearning. $AUC_{l, u}$ has risen up to $1$, indicating the discriminator outputs relatively low values for the unlearning images compared with the learning images. 
However, the involvement of the perceptual loss downgrades the model's intrinsic performance and downstream task. When $\lambda_2$ changes from $0$ to $1$, $FID_{l}$ gets larger, almost twice when unlearning $64$ and $1024$ images and even three times when unlearning $256$ images. It indicates significant degradation in the model's intrinsic performance. The classification accuracy $ACC$ also suffers from degradation. For example, when unlearning $256$ images, the $ACC$ is $0.78$, $0.72$, and $0.67$ separately for the case of $\lambda_2$ being $0$, $0.5$, and $1$. 
The only advantage is the improvement in efficiency. An unlearning with a larger $\lambda_2$ spends less time to finish the unlearning, however, the acceleration shows up only for the case of unlearning $1024$ images. 

Under comprehensive consideration, we conclude that the few-shot cascaded unlearning can finish unlearning by consuming less time, however, inducing severe degradation in the model's intrinsic performance and downstream task. Hence, we recommend no perceptual distance i.e., $\lambda_2=0$ for item unlearning.

\paragraph{class unlearning}

Table \ref{tab:class unlearning of CIFAR perceptual loss} displays the results of class unlearning on CIFAR-$10$. When unlearning a class, the perceptual distance plays a successful role in accelerating the few-shot cascaded unlearning. Take the \textit{average} substitute mechanism as an example. The unlearning needs at least $35060.71$ seconds to reach the destination when $\lambda_2$ is $0$, whereas only needs $4414.34$ and $3089.21$ seconds for the cases where $\lambda_2$ is $0.5$ and $1$. This acceleration, however, comes at the cost of a reduction in both the model's intrinsic performance and its performance on the classification task, particularly evident with $\lambda_2=1$. The $FID_{l}$ value doubles, and the accuracy $ACC$ drops to as low as $0.70$, compared to $0.75$ with $\lambda_2=0$. Figure \ref{fig: unlearning class of CIFAR-$10$} presents the generated images of the unlearning and other labels. When $\lambda_2=1$, we can observe less fidelity from  the generated images of the other labels. In comparison, $\lambda_2=0.5$ is a compromise, where both the model's intrinsic performance and downstream task performance experience only slight degradation, marked by a modest increase in $FID_{l}$ and a decrease in $ACC$, all while achieving a significant time reduction of at least $6.9$ times. There might be more suitable values for $\lambda_2$. It's worth noting that there may exist more suitable values for $\lambda_2$. Striking a balance, $\lambda_2$ can be set in a manner that prevents a significant increase in $FID_{l}$ and decrease in $ACC$, all while considerably reducing the time required. Our future work will delve further into exploring these possibilities.

% Table generated by Excel2LaTeX from sheet 'Sheet5'
\begin{table*}[htbp]
  \centering
  \caption{Statistical results with changing $\lambda_2$ on CIFAR-$10$. The cascaded unlearning unlearns $64$, $256$, and $1024$ images with $F_{label}=0.1$, $\lambda_1=1$, and $\lambda_2=[0,0.5,1]$.}
    \begin{tabular}{lrrrrrrrrr}
    \hline
    unlearn num & \multicolumn{3}{c}{64 } & \multicolumn{3}{c}{256 } & \multicolumn{3}{c}{1024 } \\
    $\lambda_2$ & \multicolumn{1}{l}{0} & \multicolumn{1}{l}{0.5} & \multicolumn{1}{l}{1} & \multicolumn{1}{l}{0} & \multicolumn{1}{l}{0.5} & \multicolumn{1}{l}{1} & \multicolumn{1}{l}{0} & \multicolumn{1}{l}{0.5} & \multicolumn{1}{l}{1} \\
    \hline
    \rowcolor{lightgray}
    pre $AUC_{l,u}$ & 0.49  & 0.49  & 0.49  & 0.44  & 0.44  & 0.44  & 0.49  & 0.49  & 0.49  \\
    $AUC_{l,u}$ & 0.98  & 0.98  & 0.97  & 0.98  & 0.96  & 0.99  & 1.00  & 0.99  & 0.99  \\
    \rowcolor{lightgray}
    pre $FID_{l}$ & 5.35  & 5.35  & 5.35  & 5.52  & 5.52  & 5.52  & 5.48  & 5.48  & 5.48  \\
    $FID_{l}$ & 7.80  & 12.21  & 15.58  & 9.03  & 20.35  & 27.24  & 9.51  & 16.52  & 19.75  \\
    \rowcolor{lightgray}
    pre $ACC$ & 0.75  & 0.75  & 0.75  & 0.77  & 0.77  & 0.77  & 0.76  & 0.76  & 0.76  \\
    $ACC$ & 0.77  & 0.73  & 0.72  & 0.77  & 0.69  & 0.65  & 0.74  & 0.67  & 0.65  \\
    $T$ & 496.65  & 495.39  & 499.97  & 1534.39  & 1488.31  & 1549.77  & 14392.24  & 12968.21  & 11566.69  \\
    \hline
    \end{tabular}%
  \label{tab:item unlearning of CIFAR perceptual loss}%
\end{table*}%

% Table generated by Excel2LaTeX from sheet 'Sheet5'
\begin{table}[htbp]
  \centering
  \caption{Statistical results with changing $\lambda_2$ on CIFAR-$10$. The cascaded unlearning unlearns the class whose label is $7$ with $F_{label}=0.1$, $\lambda_1=1$, and $\lambda_2=[0,0.5,1]$.}
  \resizebox*{0.5\textwidth}{!}{
    \begin{tabular}{lcccccc}
    \hline
    substitute & \multicolumn{3}{c}{average} & \multicolumn{3}{c}{other-class} \\
    $\lambda_2$ & \multicolumn{1}{c}{0} & \multicolumn{1}{c}{0.5} & \multicolumn{1}{c}{1} & \multicolumn{1}{c}{0} & \multicolumn{1}{c}{0.5} & \multicolumn{1}{c}{1} \\
    \hline
    \rowcolor{lightgray}
    pre $FID_{u}$ & 12.47  & 12.47  & 12.47  & 12.47  & 12.47  & 12.47  \\
    $FID_{u}$ & 152.02  & 161.63  & 164.08  & 150.25  & 160.92  & 177.89  \\
    \rowcolor{lightgray}
    pre $AUC_{l,u}$ & 0.62  & 0.62  & 0.62  & 0.62  & 0.62  & 0.62  \\
    $AUC_{l,u}$ & 1.00  & 1.00  & 1.00  & 1.00  & 1.00  & 0.79  \\
    \rowcolor{lightgray}
    pre $FID_{l}$ & 5.60  & 5.61  & 5.60  & 5.60  & 5.60  & 5.60  \\
    $FID_{l}$ & 10.87  & 13.87  & 21.20  & 13.79  & 14.62  & 25.93  \\
    \rowcolor{lightgray}
    pre $ACC$ & 0.78  & 0.78  & 0.78  & 0.78  & 0.78  & 0.78  \\
    $ACC$ & 0.75  & 0.75  & 0.70  & 0.73  & 0.77  & 0.72  \\
    $T$ & 35060.72  & 4416.34  & 3089.21  & 64304.34  & 3570.07  & 3208.45  \\
    \hline
    \end{tabular}%
    }
  \label{tab:class unlearning of CIFAR perceptual loss}%
\end{table}%

\begin{figure}[htbp]
    \centering
    % \subfigure[Original unlearn]{
    %     \includegraphics[scale=0.42]{images/exp/cifar-class-image/17unlearn_data_average_0.1_50_0_0.00/0_0_img_gen_unlearn.jpg}}
    % \subfigure[Original learn]{
    %     \includegraphics[scale=0.42]{images/exp/cifar-class-image/17unlearn_data_average_0.1_50_0_0.00/0_0_img_gen_learn.jpg}}
    \subfigure[Few-shot (average, $\lambda_2=0.5$)]{
    \includegraphics[scale=0.42]{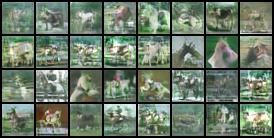}}
    \subfigure[Few-shot (average, $\lambda_2=0.5$)]{
    \includegraphics[scale=0.42]{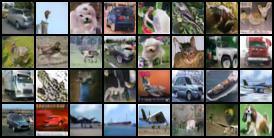}}
    \subfigure[Few-shot (average, $\lambda_2=1$)]{
    \includegraphics[scale=0.42]{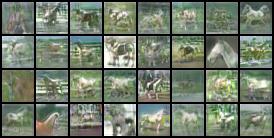}}
    \subfigure[Few-shot (average, $\lambda_2=1$)]{
    \includegraphics[scale=0.42]{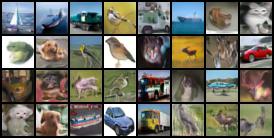}}
    \subfigure[Few-shot (other-class, $\lambda_2=0.5$)]{
    \includegraphics[scale=0.42]{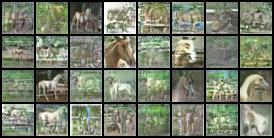}}
    \subfigure[Few-shot (other-class, $\lambda_2=0.5$)]{
    \includegraphics[scale=0.42]{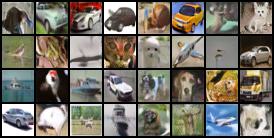}}
    \subfigure[Few-shot (other-class, $\lambda_2=1$)]{
    \includegraphics[scale=0.42]{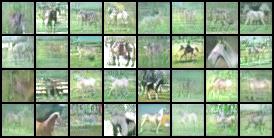}}
    \subfigure[Few-shot (other-class, $\lambda_2=1$)]{
    \includegraphics[scale=0.42]{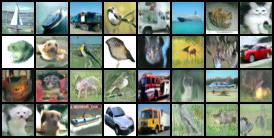}}
    \caption{Visual results of class unlearning on CIFAR-$10$ dataset with few-shot cascaded unlearning which involves perceptual loss, and the unlearning label index is $7$. The left column is of unlearning label i.e., $7$ and the right column is of the other labels. And the first row comes from the target GAN without unlearning operations. $F_{label}$=0.1, $\lambda_1=1$}
    \label{fig: unlearning class of CIFAR-$10$: perceptual loss}
\end{figure}

\subsection{Inference evaluation}
\label{sec:Inference evaluation}
A membership inference attack (MIA) is a popular unlearning measurement in classifier unlearning works \cite{nguyenSurveyMachineUnlearning2022, DBLP:conf/ccs/Chen000HZ21}. This kind of attack aims to detect whether an image is used to train a target model. In the scenario of unlearning measurement, for an unlearning image, if the MIA detects it as a member before unlearning and subsequently as a non-member after unlearning, we conclude the target model has unlearned the unlearning image, in other words, successfully finishing the unlearning. We select a discriminator-based attack, i.e., LOGAN attack \cite{hayesLOGANMembershipInference2019} as the membership inference tool because the discriminator has direct access to the training images, thus deemed most effective. 

LOGAN attack utilizes the overfitting discriminator $D$ which places a higher confidence value on members of the training dataset. To evaluate the attack, we introduce the test dataset $D_{t}$ as non-members which follow the distribution of the training dataset but are not used to train the target GAN. According to the statement LOGAN, with a successful unlearning, $D(x)_{x\sim D_u}$ is higher than $D(x)_{x\sim D_t}$ before unlearning and gets lower after unlearning. We use the AUC score for the comparison, $AUC_{u,t} = P(D(x)_{x\sim D_u}>D(x)_{x\sim D_t})$.
Within a class unlearning, the test images of $D_t$ have the same label as the unlearning class, under the consideration that the model performs differently on images of different labels. If $AUC_{u,t}$ decreases, the risk of membership inference is mitigated and the unlearning is successful to some extent.
Empirically, we conduct the LOGAN attack on item and class unlearning tasks in section \ref{sec:Item unlearning}, including few-shot and zero-shot cascaded unlearning on MNIST and CIFAR-$10$ datasets.

% \paragraph{MNIST}
Table \ref{tab:LOGAN MIA on MNIST}, \ref{tab:LOGAN MIA on CIFAR-$10$}, and \ref{tab:LOGAN MIA on FFHQ} depict the LOGAN results of the target GAN model on MNIST, CIFAR-$10$, and FFHQ before and after our cascaded unlearning. $AUC_{u, t}$ decreases to round or below $0.5$ for all cases overall, but there are still the following differences:
$1.$ As for item unlearning of MNIST, we observe significant decreases in $AUC_{u, t}$ of both few-shot and zero-shot cascaded unlearning. It indicates that $D(x)_{x\sim D_u}$ values of unlearning images decrease much and are smaller than $D(x)_{x\sim D_t}$ values of test images.
$2.$ As for item unlearning of CIFAR-$10$, we also observe the decreases in $AUC_{u, t}$, whereas $AUC_{u, t}$ decreases less when unlearns more in the few-shot unlearning. $AUC_{u, t}$ values are still under $0.55$.
% We attribute it to the involvement of learning images which prevent over-unlearning as well as unlearning. 
$3.$ The decreases in class unlearning of MNIST and CIFAR-$10$ are also moderate, $AUC_{u, t}$ from $0.548$ to around $0.5$ for MNIST and from $0.615$ to around $0.5$. 

Actually, we only expect the discriminator to have an equivalent or poorer perception of the unlearning images compared with the test images, that is to say, $AUC_{u, t}$ is around or less than $0.5$ is acceptable. According to such $AUC_{u, t}$ values, we cannot decide which image is a member with high confidence. Hence, we conclude the few-shot and zero-shot cascaded unlearning bypass the LOGAN inference on MNIST and CIFAR-$10$ datasets, as well as the few-shot cascaded unlearning bypass the LOGAN inference on FFHQ datasets.

% As for item unlearning, we observe significant decreases in $AUC_{u, t}$ of both few-shot and zero-shot cascaded unlearning. It indicates that $D(x)_{x\sim D_u}$ values of unlearning images decrease much and are smaller than $D(x)_{x\sim D_t}$ values of test images. In other words, the GAN model has a poorer perception of the unlearning images than of the test images. We cannot deduce membership through overfitting anymore. Additionally, $AUC_{u, t}$ values of the few-shot cascaded unlearning are a bit higher, suggesting the involvement of learning images possibly prevents the unlearning as it prevents over-unlearning. 

% By comparison, the decreases in class unlearning are moderate, $AUC_{u, t}$ from $0.548$ to around $0.5$. We conclude the few-shot and zero-shot cascaded unlearning bypass the LOGAN inference because we cannot decide which image is a member with a higher possibility according to $AUC_{u, t}$. Actually, we only expect the discriminator to have an equivalent or poorer perception of the unlearning images compared with the test images, that is to say, $AUC_{u, t}$ is around or less than $0.5$ is acceptable. 

% Table generated by Excel2LaTeX from sheet 'mnist (2)'
\begin{table}[htbp]
  \centering
  \caption{LOGAN MIA on the GAN model (MNIST) before and after our cascaded unlearning. The settings of the cascaded unlearning are $F_{label}=0.1$, $\lambda_1=1$, and $\lambda_2=0$. A higher $AUC_{u,t}$ denotes a higher possibility of correct inferences.}
  \resizebox*{0.5\textwidth}{!}{
    \begin{tabular}{ccccc}
    \hline
    \multirow{2}[0]{*}{unlearn} & \multicolumn{1}{c}{\multirow{2}[0]{*}{substitute}} & \multirow{2}[0]{*}{pre $AUC_{u, t}$} & \multicolumn{2}{c}{$AUC_{u, t}$} \\
          & \multicolumn{1}{c}{} &       & Few-shot & zero-shot \\
    \hline
    \multirow{3}[0]{*}{item-64} & average & 0.634  & 0.208  & 0.167  \\
          & projection & 0.634  & 0.186  & 0.148  \\
          & truncation & 0.634  & 0.191  & 0.154  \\
    \multicolumn{1}{c}{\multirow{3}[0]{*}{item-256}} & average & 0.620  & 0.196  & 0.187  \\
          & projection & 0.620  & 0.230  & 0.159  \\
          & truncation & 0.620  & 0.206  & 0.185  \\
    \multicolumn{1}{c}{\multirow{3}[0]{*}{item-1024}} & average & 0.583  & 0.224  & 0.199  \\
          & projection & 0.583  & 0.219  & 0.153  \\
          & truncation & 0.583  & 0.221  & 0.200  \\
    \multicolumn{1}{c}{\multirow{2}[0]{*}{class}} & average & 0.548  & 0.482  & 0.501  \\
          & \shortstack{other-class} & 0.548  & 0.487  & 0.522  \\
    \hline
    \end{tabular}%
    }
  \label{tab:LOGAN MIA on MNIST}%
\end{table}%

% \paragraph{CIFAR-$10$}
% Table \ref{tab:LOGAN MIA on CIFAR-$10$} depicts the LOGAN results of the target GAN model on CIFAR-$10$ before and after our cascaded unlearning. We also observed decreases in both item and class unlearning of both few-shot and zero-shot cascaded unlearning. All of $AUC_{u, t}$ values are under $0.55$, therefore, we conclude our few-shot and zero-shot cascaded unlearning 

% Table generated by Excel2LaTeX from sheet 'cifar (2)'
\begin{table}[htbp]
  \centering
  \caption{LOGAN MIA on the GAN model (CIFAR-$10$) before and after our cascaded unlearning. The settings of the cascaded unlearning are $F_{label}=0.1$, $\lambda_1=1$, and $\lambda_2=0$. A higher $AUC_{u,t}$ denotes a higher possibility of correct inferences.}
  \resizebox*{0.5\textwidth}{!}{
    \begin{tabular}{ccccc}
    \hline
    \multirow{2}[0]{*}{unlearn} & \multicolumn{1}{c}{\multirow{2}[0]{*}{substitute}} & \multirow{2}[0]{*}{pre $AUC_{u, t}$} & \multicolumn{2}{c}{$AUC_{u, t}$} \\
          & \multicolumn{1}{c}{} &       & Few-shot & zero-shot \\
    \hline
    \multirow{3}[0]{*}{item-64} & average & 0.601  & 0.169  & 0.205  \\
          & projection & 0.601  & 0.182  & 0.205  \\
          & truncation & 0.601  & 0.169  & 0.203  \\
    \multicolumn{1}{c}{\multirow{3}[0]{*}{item-256}} & average & 0.618  & 0.547  & 0.158  \\
          & projection & 0.618  & 0.523  & 0.158  \\
          & truncation & 0.618  & 0.526  & 0.174  \\
    \multicolumn{1}{c}{\multirow{3}[0]{*}{item-1024}} & average & 0.599  & 0.537  & 0.184  \\
          & projection & 0.599  & 0.537  & 0.209  \\
          & truncation & 0.599  & 0.456  & 0.166  \\
    \multicolumn{1}{c}{\multirow{2}[0]{*}{class}} & average & 0.615  & 0.440  & 0.407  \\
          & \shortstack{other-class} & 0.615  & 0.469  & 0.466  \\
    \hline
    \end{tabular}%
    }
  \label{tab:LOGAN MIA on CIFAR-$10$}%
\end{table}%

% Table generated by Excel2LaTeX from sheet 'Sheet (2)'
\begin{table}[htbp]
  \centering
  \caption{LOGAN MIA on the GAN model (FFHQ) before and after our cascaded unlearning. Using few-shot and \textit{average} substitute mechanism with $F_{label}$=0.1, $\lambda_1=1$, and $\lambda_2=0$. A higher $AUC_{u,t}$ denotes a higher possibility of correct inferences.}
    \begin{tabular}{p{6.5em}cc}
    \hline
    \multicolumn{1}{l}{unlearn num} & \multicolumn{1}{l}{pre $AUC_{u,t}$} & \multicolumn{1}{l}{$AUC_{u,t}$} \\
    \hline
    64 & 0.58  & 0.42  \\
    256   & 0.68  & 0.55  \\
    1024  & 0.69  & 0.50  \\
    \hline
    \end{tabular}%
  \label{tab:LOGAN MIA on FFHQ}%
\end{table}%

\subsection{Comparison with the baseline}
\begin{table*}[htbp]
  \centering
  \caption{Comparison with the baseline method i.e., retraining. Lower $FID_{l}$, higher $ACC$, and smaller $T$ respectively indicate better performances in model self, downstream task, and unlearning efficiency. The column \textit{Saving times} records how many times the cascaded unlearning saves time.}
  \renewcommand\arraystretch{1.3}
    \begin{tabular}{cccccrrrrc}
    \hline
    \multirow{2}[2]{*}{Dataset} & \multirow{2}[2]{*}{Unlearn} & \multicolumn{3}{c}{Retraining} & \multicolumn{5}{c}{Cascaded unlearning} \\
          &       & $FID_{l}$   & $ACC$ & $T$  & \multicolumn{1}{c}{$FID_{l}$} & \multicolumn{1}{c}{$ACC$} & \multicolumn{1}{c}{$T$} & \multicolumn{1}{l}{Saving times} & Parameters \\
    \hline
    \multirow{4}[1]{*}{MNIST} & item-64 & 3.28  & 0.80  & 66847.64  & 3.43  & 0.83  & 360.84  & \textbf{185.26}  & \shortstack{Few-shot, average,\\$F_{label}$=0.1,$\lambda_1=1$, $\lambda_2=0$} \\
          & item-256 & 2.22  & 0.82  & 70967.10  & 2.34  & 0.82  & 2026.99  & 35.01  & \shortstack{Few-shot, average,\\$F_{label}=0.1$,$\lambda_1=1$, $\lambda_2=0$} \\
          & item-1024 & 1.60  & 0.81  & 67762.38  & 4.64  & 0.82  & 13689.33  & 4.95  & \shortstack{Few-shot, average,\\$F_{label}=0.1$,$\lambda_1=1$, $\lambda_2=0$} \\
          & class & 3.26  & 0.79  & 70833.25  & 3.97  & 0.85  & 7674.45  & 9.23  & \shortstack{Few-shot, average,\\$F_{label}=0.1$,$\lambda_1=1$, $\lambda_2=0$} \\
    \hline
    \multirow{4}[1]{*}{CIFAR-$10$} & item-64 & 5.54  & 0.69  & 142910.58  & 7.84  & 0.77  & 502.56  & \textbf{284.37}  & \shortstack{Few-shot, average,\\$F_{label}=0.1$,$\lambda_1=1$, $\lambda_2=0$} \\
          & item-256 & 5.50  & 0.72  & 138572.69  & 8.98  & 0.77  & 1530.59  & 90.54  & \shortstack{Few-shot, average,\\$F_{label}=0.1$,$\lambda_1=1$, $\lambda_2=0$} \\
          & item-1024 & 7.11  & 0.68  & 69956.67  & 9.51  & 0.74  & 12978.90  & 5.39  & \shortstack{Few-shot, average,\\$F_{label}=0.1$,$\lambda_1=1$, $\lambda_2=0$} \\
          & class & 6.26  & 0.72  & 143265.23  & 13.87  & 0.75  & 4416.34  & 32.44  & \shortstack{Few-shot, average,\\$F_{label}=0.1$,$\lambda_1=1$, $\lambda_2=0.5$} \\
    \hline
    \end{tabular}%
  \label{tab:baseline performance}%
\end{table*}%

In this section, we conduct retraining as the exact unlearning. The results are shown in table \ref{tab:baseline performance}. We do not discuss the unlearning effect because retraining definitely provides the perfect unlearning. In other words, we compare our cascaded unlearning with the retraining from three aspects: the model's intrinsic performance, the downstream task such as the classification, and consuming time. We conclude as follows:
\begin{enumerate}[leftmargin=*]
    \item For MNIST, we do not observe any significant differences in the model's intrinsic performance $FID$ and downstream task $ACC$ except for unlearning $1024$ images. When unlearning $1024$ images, $FID_{l}$ of the cascaded unlearning is higher, however, $ACC$ is equally high.
    \item For CIFAR-$10$, the cascaded unlearning invokes a degradation in the model's intrinsic performance with higher $FID_{l}$. The more images to unlearn, the wider differences of $FID_{l}$ between the cascaded unlearning and retraining becomes. When unlearning $64$ images, $FID_{l}$ of retraining is $5.54$ while $FID_{l}$ of the cascaded unlearning is $7.84$.
    \item The cascaded unlearning has higher classification accuracy $ACC$ values than retraining for both MNIST and CIFAR-$10$.
    \item The cascaded unlearning definitely saves much time for both MNIST and CIFAR-$10$, especially for the scenario of unlearning $64$ images. For example, when unlearning $64$ images of MNIST, the cascaded unlearning spends $360.84$ seconds, while retraining spends $66847.64$ seconds, which is $185$ times more than the cascaded unlearning. However, the advantage has been weakened as more images to unlearn. To keep the advantage, we apply perceptual distance in the cascaded unlearning with $\lambda_2=0.5$, which spends one-thirty-second of the retraining's time.
\end{enumerate}
Finally, it's worth noting that cascaded unlearning offers enhanced privacy preservation. Traditional retraining requires all learning images, whereas cascaded unlearning only necessitates equivalent learning images for the few-shot case or no learning images at all for the zero-shot case. This distinction holds significant value in preventing the exposure of training images.

% % Table generated by Excel2LaTeX from sheet 'Sheet1 (2)'
% \begin{table}[htbp]
%   \centering
%   \caption{Baseline performance}
%     \begin{tabular}{ccccc}
%     \hline
%      & Unlearn & FID   & Accuracy & Time \\
%      \hline
%     \multirow{4}[0]{*}{MNIST} & item-64 & 3.28  & 0.80  & 66847.64  \\
%           & item-256 & 2.22  & 0.82  & 70967.10  \\
%           & item-1024 & 1.60  & 0.81  & 67762.38  \\
%           & class & 3.26  & 0.79  & 70833.25  \\
%     \hline
%     \multirow{4}[0]{*}{CIFAR-$10$} & item-64 & 5.54  & 0.69  & 142910.58  \\
%           & item-256 & 5.50  & 0.72  & 138572.69  \\
%           & item-1024 & 7.11  & 0.68  & 69956.67  \\
%           & class & 6.26  & 0.72  & 143265.23  \\
%     \hline
%     \end{tabular}%
%   \label{tab:baseline performance}%
% \end{table}%

\section{Conclusion}

This paper proposes a cascaded unlearning to achieve GAN unlearning, erasing several specific images or a class from a trained GAN model. We first introduce a substitute mechanism and a fake label to tackle the unlearning challenges in the generator and discriminator, which comprise a GAN model. Then we propose a cascaded unlearning in a few-shot case and a more privacy-preserving zero-shot case. We establish a comprehensive evaluation framework encompassing four key dimensions: the unlearning effectiveness, the intrinsic performance of the model, the performance of downstream tasks like classification, and unlearning efficiency. Experimental results indicate the cascaded unlearning can efficiently finish item and class unlearning with no significant degradation in the performances of the model itself and downstream tasks such as classification.

While the cascaded unlearning algorithm has made valuable contributions, there are several fields for future research that merit further exploration and improvement. One notable challenge is extending the algorithm's capabilities to handle large-scale datasets, particularly high-quality images such as those with dimensions of $1024\times1024\times3$. The experimental results presented in this paper indicate that unlearning minor images can diminish model fidelity for high-quality datasets like FFHQ ($256\times256\times3$). Given the prevalence of GAN models in high-quality image generation and editing, the endeavor to scale our cascaded unlearning approach to high-quality datasets holds significant significance.
Moreover, enhancing the privacy-preserving aspects of the cascaded unlearning process is a practical necessity. Although the cascaded unlearning requires few or no remaining training images, it necessitates all the unlearning images. Exploring ways to potentially reduce the demand for unlearning images, particularly in cases of class unlearning, could be an avenue for further investigation.
Research in the realm of GAN unlearning is still in its nascent stages, and more attention needs to be paid to encourage more effective and practical methodologies.

%notably, consuming less time. For example, when unlearning a minimum of images such as $64$, it improves time to unlearn points from the MNIST dataset by $185$x, and $284$x for the CIFAR-$10$ dataset, over retraining from scratch, with a slightly poorer model performance and equivalent classification accuracy.

\bibliographystyle{IEEEtran}
\bibliography{bib}

\end{document}